\documentclass[sigconf,nonacm]{acmart}

\setcopyright{none}

\usepackage{amsmath}
\usepackage{booktabs}
\usepackage{graphicx}
\usepackage{tikz}
\usetikzlibrary{positioning,arrows.meta,fit,backgrounds,calc,shapes.geometric}
\usepackage{xcolor}

\hypersetup{
  pdfauthor={Xuanyi Li, Vaskar Nath, Hossein Amirkhani, Jay Li, Alex Deng},
  pdftitle={From Offline Proxies to Online Decisions: A Layered Engagement Evaluation Framework for Conversational AI},
  pdfsubject={},
  pdfkeywords={offline evaluation, online experiments, conversational AI}
}
\ifdefined\pdfinfoomitdate\pdfinfoomitdate=1\fi
\ifdefined\pdftrailerid\pdftrailerid{}\fi

\begin{document}

\title{From Offline Proxies to Online Decisions: A Layered Engagement Evaluation Framework for Conversational AI}

\author{Xuanyi Li}
\affiliation{%
  \institution{Meta Platforms, Inc.}
  \city{Menlo Park}
  \state{California}
  \country{USA}
}
\email{xuanyili@meta.com}

\author{Vaskar Nath}
\affiliation{%
  \institution{Meta Platforms, Inc.}
  \city{Menlo Park}
  \state{California}
  \country{USA}
}
\email{vaskarnath@meta.com}

\author{Hossein Amirkhani}
\affiliation{%
  \institution{Meta Platforms, Inc.}
  \city{Menlo Park}
  \state{California}
  \country{USA}
}
\email{hossein@meta.com}

\author{Jay Li}
\affiliation{%
  \institution{Meta Platforms, Inc.}
  \city{Menlo Park}
  \state{California}
  \country{USA}
}
\email{jayl@meta.com}

\author{Alex Deng}
\affiliation{%
  \institution{Meta Platforms, Inc.}
  \city{Menlo Park}
  \state{California}
  \country{USA}
}
\email{adeng@meta.com}

\renewcommand{\shortauthors}{Li et al.}

\begin{abstract}
Online A/B experiments are the decision standard for user engagement, but traffic
and readout time limit how many conversational-AI changes can be tested. We ask
whether an offline signal designed to be computable without treatment-arm user
exposure agrees with the outcomes of those experiments.

We contribute a reusable construction and diagnosis checklist that treats an
offline proxy as a chain of three alignments---behavioral label to product
outcome, learned classifier to candidate-assistant behavior, and aggregated
offline signal to experiment effect. A companion evaluation protocol audits the whole
composite by interval-aware decision agreement, which compares offline and online confidence intervals instead of point estimates, and by within-experiment ranking. The instantiation we evaluate comprises a fixed evaluation suite on which candidate behavior
is scored, 
an engagement classifier trained to predict session/prompt level engagements, and a calibration layer mapping sample-level score differences to online model-level
engagement deltas.

We then report the audit: 489 paired offline--online contrasts (one candidate arm
against its control) from 27 experiments on a deployed multi-turn assistant,
spanning model checkpoints to system-prompt tuning. Our primary test uses the 113
contrasts from eight experiments that ran after the map was frozen: on these the
composite reaches 81.1\% F1, against 34.3\% for the raw classifier score it is
built on, and makes no wrong-direction calls where that raw score makes 31.

Every offline prediction was computed before its experiment ran to prevent overfitting. The evidence supports using the composite to prioritize candidates before scarce experiment traffic is allocated---in our deployment of the experiment, selecting among training checkpoints and tuning system prompts.
\end{abstract}



\maketitle

\section{Introduction}
\label{sec:intro}
Randomized online experiments are the decision standard for whether a production
change improves user engagement, but they are capacity constrained: each test
consumes traffic, needs a multi-day readout, and competes for a limited slot.
Overlapping designs reuse traffic across many concurrent
tests~\cite{tang2010overlapping,kohavi2020trustworthy}, but only when treatments
touch independent surfaces.
In the program studied here an arm-level engagement readout takes seven days, and
a typical experiment resolves only a handful of the candidates teams would like to
compare. An offline proxy could inform prioritization, but only if its
relationship to online outcomes is measured rather than assumed. The relevant
question is whether it predicts the direction and ordering of engagement effects later observed in randomized experiments.

We study this question in an experiment involving a multiturn, tool-using
conversational AI assistant, where the measurement unit is assistant behavior
within an ongoing dialog. Continued interaction can expose failures
invisible at the current turn, yet simulating it adds its own mismatch, so we keep
the two separate: the main audit uses a fixed static suite, and
simulated continuations are a distinct suite-design extension.

Our observation is that such an offline engagement evaluation is not easily validated by one single metric. It is a chain of three alignments---the training label to the product outcome, the classifier to candidate-assistant behavior, and the suite
distribution, and aggregation to the between-arm differences. Together these form a construction and diagnosis checklist in which a failure at an earlier layer cannot be repaired by a more elaborate later one.

We instantiate the checklist with 1) an engagement classifier (EC) trained to predict
per-session engagement as a measurement instrument, 2) a fixed suite produces arm-level EC scores, 3) a calibration converts score differences into an engagement signal with an offline bootstrapped interval.

The evaluation lifecycle records, for each candidate arm relative to control, paired
observations of the form
\[
  \left(\widehat{\Delta}^{\mathrm{offline}},\, I^{\mathrm{offline}},\,
  \Delta^{\mathrm{online}},\, I^{\mathrm{online}}\right),
\]
the predicted and realized online engagement deltas with their offline
scoring-stability and confidence intervals, the hat marking the proxy's
prediction. This corpus turns offline--online agreement into an auditable
quantity.

\paragraph{Evidence boundary.}
One provenance statement governs every result. The frozen calibration map and all
offline evaluation outputs predate treatment exposure for the cohorts they are
evaluated on; versioned records establish the freeze, and offline inputs contain
no treatment-arm logs and no online outcomes. We assembled the result tables, however, from a
read-only copy of the data after the outcomes were known, and neither the analysis
plan nor the split by model tier was fixed in advance. Every claim below therefore rests on evidence
computed before exposure but analyzed afterwards, and we do not restate this at
each result.

\paragraph{Scope.}
Three chronologically disjoint slices supply 489 paired contrasts from 27
experiments: a descriptive 286-contrast slice where the frozen export and
completed online readouts overlap, which includes development
data, a 90-contrast post-freeze discovery cohort, and a 113-contrast post-freeze replication cohort. Around that spine we compare against five other deployed offline
evaluators and against less computationally expensive random-sign, majority, point-sign, and
single-signal rules, and run a component ablation, three suite-design
studies, and horizon, confidence-level, and
clustering sensitivities. The supplement indexes each slice's overlap and evidentiary
status, so supporting evidence is never read as independent replication. Across them the frozen composite reaches 82.9\% contrast-micro F1 on
the completed overlap, against 75.8\% for the raw classifier score, and 81.1\% on
the disjoint replication cohort, against 34.3\% with zero wrong-direction calls
versus 31. It exceeds the strongest of five other deployed offline evaluators by
25.7 F1 points.

The paper contributes, in order, a checklist, an evaluation method, and evidence:
\begin{itemize}
  \item \textbf{Construction and diagnosis checklist.} Decomposing
  offline-to-online prediction into label--outcome, scorer--behavior, and
  suite--experiment alignment makes the frozen composite the unit of evaluation, and gives failures an ordered diagnosis.
  \item \textbf{Evaluation method.} We formulate predictive validity as
  online-confidence-interval-aware decision agreement and within-experiment
  ranking, using offline scoring-stability intervals for abstention, reporting both
  contrast-micro F1 (pooling all contrasts) and experiment-macro F1 (weighting
  every experiment equally), and making wrong-direction calls explicit.
  \item \textbf{Sequential frozen-composite evidence.} We built an offline evaluator that predicts the outcome of randomized online experiments on session-level engagement, and validate it frozen: tested unchanged on batches of A/B tests that ran after the freeze, with no retuning or post hoc row selection. In our experiment with 113 contrasts from eight A/B tests — it reaches 81.1\% contrast-micro F1 with no wrong-direction calls, and on a shared slice it outperforms five other deployed offline evaluators by 25.7 F1 points. Agreement is strongest where candidate and control differ in model capability tier and more modest where they match. Supporting studies isolate which components carry the signal; what we validate is the composite as a whole rather than the classifier alone.
\end{itemize}

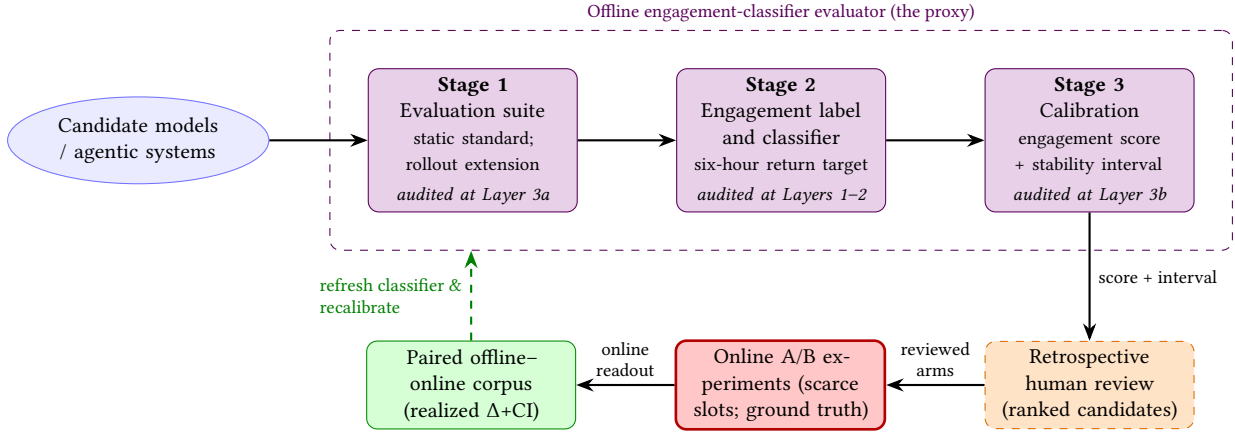
\begin{figure*}[t]
  \centering
  \begin{tikzpicture}[
      font=\small, >=Stealth, node distance=9mm and 13mm,
      box/.style={draw, rounded corners, align=center, inner sep=3pt,
        minimum height=1.15cm, text width=2.55cm},
      mlbox/.style={box, fill=violet!18, draw=violet!70!black},
      io/.style={draw, ellipse, align=center, fill=blue!8, draw=blue!55,
        inner sep=2pt, text width=2.3cm, minimum height=1.15cm},
      decision/.style={box, dashed, fill=orange!22, draw=orange!85!black},
      truth/.style={box, fill=red!22, draw=red!70!black, line width=1pt},
      data/.style={box, fill=green!16, draw=green!55!black},
      flow/.style={->, thick},
      fb/.style={->, thick, dashed, green!50!black}]
    \node[io] (cand) {Candidate models / agentic systems};
    \node[mlbox, right=of cand] (eval) {\textbf{Stage 1}\\Evaluation suite\\{\footnotesize static standard; rollout extension}\\[1pt]{\footnotesize\itshape audited at Layer 3a}};
    \node[mlbox, right=of eval] (ec) {\textbf{Stage 2}\\Engagement label and classifier\\{\footnotesize six-hour return target}\\[1pt]{\footnotesize\itshape audited at Layers 1--2}};
    \node[mlbox, right=of ec] (cal) {\textbf{Stage 3}\\Calibration\\{\footnotesize engagement score + stability interval}\\[1pt]{\footnotesize\itshape audited at Layer 3b}};
    \node[decision, below=17mm of cal] (triage) {Retrospective human review (ranked candidates)};
    \node[truth, left=of triage] (online) {Online A/B experiments (scarce slots; ground truth)};
    \node[data, left=of online] (corpus) {Paired offline--online corpus (realized $\Delta$+CI)};
    \begin{scope}[on background layer]
      \node[draw=violet!70!black, dashed, rounded corners, inner sep=5mm,
        fit=(eval)(ec)(cal),
        label={[violet!70!black,font=\footnotesize]above:Offline engagement-classifier evaluator (the proxy)}] (proxy) {};
    \end{scope}
    \draw[flow] (cand) -- (eval);
    \draw[flow] (eval) -- (ec);
    \draw[flow] (ec) -- (cal);
    \draw[flow] (cal) -- node[right,font=\footnotesize]{score + interval} (triage);
    \draw[flow] (triage) -- node[above,align=center,font=\footnotesize]{reviewed\\arms} (online);
    \draw[flow] (online) -- node[above,align=center,font=\footnotesize]{online\\readout} (corpus);
    \draw[fb] (corpus.north) -- node[left,align=left,font=\footnotesize]{refresh classifier \&\\recalibrate}
      (corpus.north|-proxy.south);
  \end{tikzpicture}
  \caption{Layered offline-to-online audit lifecycle. The offline evaluator runs
  as three pipeline stages, left to right: a suite elicits candidate behavior, the
  classifier scores it, and calibration maps arm-level score differences to an
  engagement score with a stability interval that propagates offline scoring
  variation. The checklist layers of \S\ref{sec:layers} are numbered by
  \emph{diagnosis} order---label, then scorer, then suite and calibration---not by
  execution order, which is why Layer 3 is audited at both the first and last
  stage. The main audit uses the fixed static suite, while multi-turn
  rollout is evaluated separately as a suite extension. Ranking is measured on completed
  experiments as input to human review, not as demonstrated full-funnel
  screening or traffic allocation. Subsequent randomized readouts enlarge the
  paired audit corpus used for later refresh and recalibration.}
  \label{fig:loop}
\end{figure*}

Figure~\ref{fig:loop} summarizes the lifecycle. We evaluate prioritization within
completed experiments only---not full-funnel screening or traffic savings---and
randomized experiments remain the launch evidence.

\section{Related Work}
\label{sec:related}
Online experiments remain the decision standard because they estimate effects
under randomized exposure~\cite{kohavi2009controlled,kohavi2020trustworthy}.
Surrogate-outcome research asks when less computationally expensive signals can stand in for delayed
targets and emphasizes empirical validation against those targets
~\cite{prentice1989surrogate,fleming1996surrogate,athey2023surrogate,
tripuraneni2024choosing,dimakopoulou2023evaluating}. Our proxy differs from an
early outcome observed on exposed users: it is computed from candidate behavior
before exposure and evaluated against later signed A/B effects.

\textbf{Counterfactual and recommender evaluation.} Logged-data policy evaluation
uses propensities or doubly robust estimators to correct exposure bias
~\cite{li2011unbiased,dudik2011doubly,bottou2013counterfactual,
swaminathan2015batch}. Recommender studies also test whether offline metrics
predict online performance~\cite{gilotte2018offline,krauth2020offline,
wang2023productranking,barcena2024fortune}. Our candidates have no treatment-arm
log before launch; we instead run them on a fixed suite and test whether the
resulting measurement agrees with later randomized readouts.

\textbf{Reward models and dialogue evaluators.} Reward models usually train or
rank policies from preferences~\cite{christiano2017deep,ouyang2022training,
bai2022training}, and over-optimization exposes the risk of imperfect objectives
~\cite{gao2023overoptimization}. Learned dialogue metrics and LLM judges instead
approximate response quality or human ratings~\cite{liu2016not,mehri2020usr,
ghazarian2020predictive,gao2020dialogrpt,zheng2023judging,liu2023geval,
chiang2024chatbot}. Here the classifier is neither a policy objective nor a
general-purpose judge. It is one component of a measurement instrument whose
target is agreement with randomized engagement effects; this motivates decisive-
call, wrong-direction, and within-experiment ranking metrics.

\textbf{User simulation and multi-turn measurement.} Multi-turn assistants also
change the measurement unit. Simulators support dialogue-policy evaluation
~\cite{schatzmann2007agenda,sun2021simulating}, and recent LLM user agents are
tested against real behavior~\cite{kazi2024useragents,lu2026simulate}. We make no
standalone simulator-validity claim: rollout depth and scoring are judged only by
whether they improve offline--online agreement.

\textbf{Calibration and uncertainty.} Classical probability-calibration methods
show that ranking and calibrated probabilities differ
~\cite{platt1999probabilistic,zadrozny2002transforming,guo2017calibration}. Our
calibration is instead a regression from composite offline signals to online
deltas. Bootstrap intervals propagate offline-scoring variation through the fixed
map~\cite{efron1993introduction}; they are not future-effect prediction intervals.
The distinguishing contribution is therefore the end-to-end, pre-exposure-
computable audit target rather than a new classifier, simulator, or calibration
algorithm.

\section{Methodology}
\label{sec:setup}
We first define the three alignment layers (\S\ref{sec:layers}) and the
decision-aware evaluation protocol (\S\ref{sec:metric}), then instantiate them
with an engagement scorer, evaluation suite, and calibration map
(\S\ref{sec:classifier}--\S\ref{sec:calibration}). The decomposition localizes
failures; the protocol audits the frozen composite end to end.

Each experiment compares a control with one or more assistant configurations. The
online target is a seven-day arm-level engagement delta with a confidence interval;
the scorer target is instead a binary six-hour return event. These constructs
differ in horizon, unit, aggregation, and population, so calibration tests an
empirical mapping rather than asserting equivalence. Multi-arm contrasts can share
arms and are dependent. Headline results use anonymized seven-day readouts; a
matched three-day export is used only for sensitivity analysis.

\subsection{Layered Alignment}
\label{sec:layers}
The question ``can an offline evaluator predict an online experiment?'' hides
three different validity questions:
\begin{enumerate}
  \item \textbf{Label--outcome alignment.} The behavioral label used to train the
  scorer must represent the product outcome that the online experiment measures.
  A learnable but poorly aligned label produces a precise score for the wrong
  construct.
  \item \textbf{Scorer--behavior alignment.} The learned scorer must respond to
  meaningful differences in behavior produced by candidate assistants, including
  behavior outside the logged-policy distribution. Good held-out label accuracy
  alone does not establish this property.
  \item \textbf{Suite--experiment alignment.} The offline case distribution,
  interaction depth, score aggregation, and calibration must preserve the
  between-arm differences that appear in randomized online outcomes. This final
  layer is where a frozen composite can be audited for prioritization on completed
  experiments; completed-experiment agreement alone does not establish
  full-funnel screening utility.
\end{enumerate}

The numbering is a dependency order for diagnosis, not the order in which the
pipeline executes: later-stage calibration cannot repair a misaligned label or
scorer, so a credible Layer 1 is a precondition for reading Layer 2, and both are
preconditions for reading Layer 3. Layer 3 consequently spans two pipeline stages
that sit on either side of the scorer---suite construction (3a,
\S\ref{sec:evalsuite-method}) before scoring and calibration (3b,
\S\ref{sec:calibration}) after it. Once the prerequisites are credible,
disagreement localizes hypotheses to suite coverage, interaction depth,
aggregation, or calibration. Our direct evidence audits the frozen composite
output; it does not independently validate Layers 1--2. Simulated continuations
are evaluated separately because they alter Layer-3 coverage and introduce
simulator mismatch.

\subsection{Evaluation Method: Interval-Aware Decision Agreement}
\label{sec:metric}
The final layer should be evaluated according to the decisions the proxy is meant
to support. Point correlation and regression error are insufficient on their own:
they can reward accurate estimates of near-zero effects while hiding abstention,
missed significant movers, or confident wrong-direction calls. We therefore
evaluate the proxy in two complementary modes: significant-mover classification
for deciding which arms warrant attention, and within-experiment tail ranking for
allocating scarce online slots.

For significant-mover classification, we use online uncertainty to define the
realized decision and offline repeated-scoring variation to govern abstention. For a
contrast $i$, let $I^{\mathrm{online}}_i=[L^{\mathrm{online}}_i,U^{\mathrm{online}}_i]$ be the
online confidence interval and $I^{\mathrm{offline}}_i=[L^{\mathrm{offline}}_i,U^{\mathrm{offline}}_i]$
be the offline scoring-stability interval. Define the online sign
\[
y_i =
\begin{cases}
 +1, & L^{\mathrm{online}}_i > 0,\\
 -1, & U^{\mathrm{online}}_i < 0,\\
 0, & \text{otherwise,}
\end{cases}
\]
and define the predicted sign $p_i$ analogously from $I^{\mathrm{offline}}_i$. A
decisive offline call is correct when $p_i \ne 0$ and $p_i=y_i$. It is a
wrong-direction call when $p_i y_i=-1$. If the offline interval excludes zero but
the online interval overlaps zero, the call is a precision error; if the online
interval excludes zero but the offline interval overlaps zero, the contrast is a
recall miss.

``Interval-aware'' denotes that online labels and ranking forgiveness come from
online confidence intervals; the offline stability interval, formed by
bootstrapping scoring inputs through a fixed map, is not a prediction interval for
a future online effect. Every F1 we report is therefore measured at a
single decision threshold---the abstention rule above---rather than summarized
over all thresholds as an area under a curve.

Our primary directional metrics are hard precision, recall, and F1:
\[
\mathrm{precision}
= \frac{\#\{i: p_i \ne 0 \land p_i=y_i\}}{\#\{i:p_i \ne 0\}},\qquad
\]
\[
\mathrm{recall}
= \frac{\#\{i: p_i \ne 0 \land p_i=y_i\}}{\#\{i:y_i \ne 0\}}.
\]
Let $C_e$, $P_e$, and $O_e$ be experiment $e$'s correct decisive calls, decisive
offline calls, and online-significant contrasts. Contrast-micro F1 pools these
counts over all $E$ experiments, $2\sum_e C_e/\sum_e (P_e+O_e)$, so experiments
with more contrasts weigh more; experiment-macro F1 instead averages the
per-experiment $2C_e/(P_e+O_e)$, scoring an experiment zero when it has no
decisive or significant contrasts, so every experiment weighs equally. Both treat
online outcomes unresolved at the chosen confidence level as negative decisions,
without asserting that their true effects are null.

The hard rule above scores each
predicted-significant call as fully correct or fully wrong by the online
\emph{sign} $y_i$. This over-penalizes a common, benign case: the offline proxy is
more sensitive and flags a mover, the online point estimate moves in the same
direction, but the online interval is just barely not significant. We therefore
also report a \emph{power-aware} (CI-aware) precision that gives partial credit
proportional to the online support in the predicted direction. For a
predicted-significant contrast with $p_i=+1$ and online interval
$I^{\mathrm{on}}_i=[L^{\mathrm{on}}_i,U^{\mathrm{on}}_i]$, define the support
\[
s_i = \mathrm{clip}_{[0,1]}\!\left(
  \frac{U^{\mathrm{on}}_i}{U^{\mathrm{on}}_i - L^{\mathrm{on}}_i}\right),
\]
the fraction of the online interval lying on the predicted (positive) side;
$s_i=1$ when the interval excludes zero in the right direction and $s_i=0$ when it
lies entirely in the wrong direction. The symmetric quantity
$s_i=\mathrm{clip}_{[0,1]}\!\left(-L^{\mathrm{on}}_i/(U^{\mathrm{on}}_i-L^{\mathrm{on}}_i)\right)$
applies for $p_i=-1$. Power-aware precision replaces the indicator
$\mathbf{1}[p_i=y_i]$ in the numerator with $s_i$; the hard CI-aware precision is
the special case that thresholds $s_i$ at $1$. The metric stays
conservative---wrong-direction calls still earn zero---while not punishing
near-significant agreement as if it were a sign error.

For ranking, we reconstruct per-arm offline and online scores within each
experiment from the pairwise contrasts by least squares, with the mean arm score
pinned to zero. For an experiment with $m$ arms, the top and bottom tail sizes are
$k=\min(3,\lceil 0.2m\rceil)$. We report exact precision@20\% for the top tail
(candidate winners) and bottom tail (candidate regressions) as the primary ranking
metric. As a sensitivity analysis, we also forgive a boundary swap when the
corresponding online pairwise confidence interval overlaps zero. The latter avoids
penalizing an ordering that the online readout cannot resolve, but it is not the
headline ranking result.

\subsection{Engagement Target and Scorer}
\label{sec:classifier}
The full offline instrument has four parts: the engagement classifier, the
evaluation suite it scores, the aggregation rule, and the calibration layer.
This subsection instantiates Layers 1 and 2---the behavioral label and the
scorer trained on it---while \S\ref{sec:evalsuite-method} and
\S\ref{sec:calibration} instantiate the two halves of Layer 3.

Layer 1 fixes the behavioral label, and any bounded return- or
continuation-style signal could occupy the same slot; the instantiation we audit
uses a binary target of return-session occurrence within $X$ hours, with $X=6$
(360 minutes) throughout. For a logged interaction $j$ with conversation context $c_j$ and
observed assistant response $r_j$, $y_j=1$ when the production user returns to a
new prompt session within $X$ hours of the focal message/thread session ending,
and training rows are restricted to examples with a complete $X$-hour
observation window. An LLM encoder with a binary classification head produces the
positive-class score $s_j=\sigma(g_\theta(c_j,r_j))$, trained with binary
cross-entropy. This bounded target is an observed session-level association, not a
causal claim about each return.

At evaluation time the return label is unobserved. We substitute a candidate
assistant response for $r_j$, average $s_j$ within each candidate arm, and
contrast it against control; Layer~3b calibrates those aggregate differences to
online engagement deltas. So $s_j$ is comparative rather than a calibrated
probability, causal effect, or online lift. Because the logged distribution is
policy-dependent, we test scorer, suite, aggregation, and calibration end to end,
and do not separately establish label validity or scorer generalization.

\subsection{Evaluation-Suite Design}
\label{sec:evalsuite-method}
Layer 3 begins before the scorer runs, with the distribution of cases that
candidates are asked to handle.
The headline static suite uses 4{,}832 fixed examples per arm/profile with no
simulated continuation; its source suite and scoring profile are versioned and
held fixed for every headline comparison, with internal identifiers withheld.
A naive offline suite would uniformly sample historical
production conversations, truncate each at a user prompt, and query candidates
with the preceding conversation as context. That leaves three design questions:
whether targeted sampling beats uniform sampling, which turn to cut at, and
whether to evaluate behavior beyond that turn.

\paragraph{Source-pool curation.}
From a broad random pool we retain variation in use case, language, conversation
depth, and objective complexity; remove duplicates and
templated or low-quality conversations; and exclude single-turn cases, for which
turn selection is undefined. Shallow two--three-turn conversations are
downweighted rather than dropped, so that targeting critical events does not
collapse the suite onto long dialogs alone.

\paragraph{Critical-turn selection.}
An LLM judge reads the complete logged conversation and rates each user turn by
whether the next assistant response can materially change the user's progress and
whether weaker and stronger assistants would plausibly diverge there. Criticality
thus measures evaluation leverage, not the quality of the logged response: turns
requiring context use, ambiguity resolution, derailment repair, planning, or
factual grounding rate high, whereas greetings and trivial requests do not. For
each turn rating at least three on a five-point rubric, the judge also records
discrimination potential, context dependence, evidence quality, information gain,
subsequent evidence, and whether the turn is ``make-or-break''; their product,
weighted extra for make-or-break turns, is a critical-density score. The densest
prompt becomes the cutoff, and we discard its logged response before querying
candidates. Ranking conversations by their five densest turns gives a fixed
ordering whose prefixes define the nested top-$K$ suites; the formula and full
rubric are in the supplement. Candidate outputs and online outcomes are never
judge inputs, but because the judge sees the later logged trajectory, selection is
conditioned on post-turn evidence and may favor events legible in hindsight.

\paragraph{Rollout and feasibility.}
The candidate first answers the selected prompt; a user simulator can then
continue the interaction, exposing behavior that appears only after clarification,
repair, or a tool-dependent transition. We evaluate depths zero through eight,
aggregate scores through each depth, and separately inspect final-turn scoring.
Cases whose estimated history, prompt, and generation reserve exceed the model
input budget are filtered beforehand, with lower-ranked cases pulled in to restore
the requested top-$K$ size. These three choices---rollout depth, top-$K$ size, and
a critical-turn versus last-user-turn bundle---are exactly the design studies of
\S\ref{sec:scaling}, with token fit as a feasibility stress test. The headline
calibrated export uses the fixed static suite; rollout is a separately evaluated
extension.

\subsection{Offline-to-Online Calibration}
\label{sec:calibration}
Layer 3 closes with the map from an aggregated suite score to the online
experiment effect. Raw engagement-classifier deltas are not directly interpretable as online
engagement deltas, so we fit a calibration model using only features computable
before treatment-arm exposure. Let $x_i$ collect the offline signal, its
uncertainty summaries, and anonymized candidate or evaluation descriptors for
contrast $i$. A specification $s$ selects and transforms features from this pool,
giving the general map $\widehat{\Delta}^{\mathrm{online}}_i=f_{\theta_s}(\phi_s(x_i))$,
where both $s$ and $\theta_s$ are learned only from historical paired rows. The
retained specification is a no-intercept weighted sum of a small set of
pre-exposure meta features,
\[
  \widehat{\Delta}^{\mathrm{online}}_i
  = d_h \sum_{k=1}^{K} \beta_k\, z_{ik},
  \qquad d_3 = 1,
\]
where each $z_{ik}$ is a raw, unstandardized continuous descriptor of contrast
$i$ available before exposure---the classifier-score delta together with
auxiliary behavioral and configuration metadata---and $d_h$ is a multiplicative
readout-duration factor with three days as reference and a fitted seven-day
level. The retained map uses $K=4$ such features. It contains no intercept,
bucket or categorical encoding, experiment identity, date, organizational source,
or online outcome. All coefficients are withheld, as are the individual
feature identities apart from the tier term, on which the strata reported
throughout depend; what the structure establishes is that the retained predictor
is a weighted composite rather than a classifier-only calibration. That tier
term is coarse ordinal metadata assigned before offline scoring and before
exposure, recording only whether the candidate is above, below, or in the same
anonymized tier as its control. Because it is available before launch and drives
much of the cross-tier correction, we report the two strata separately.

\paragraph{Selecting the specification.}
We tune with a temporal split that keeps whole experiments together, fitting each
of 32 candidate specifications on earlier experiments under inverse
online-CI-width weighting, with leave-one-experiment-out predictions inside the
fitting slice to reduce same-experiment leakage. For split $a$, let
$H_{90}^{a}(s)$ be the smallest symmetric half-width around the predicted deltas
that overlaps at least 90\% of the observed online confidence intervals. We select
primarily by $\operatorname{robustH}_{90}(s)=\max\{H_{90}^{\mathrm{fit}}(s),H_{90}^{\mathrm{sel}}(s)\}$,
breaking ties by split-specific uncertainty widths and online-delta error, then
refit the selected form on the historical corpus. Robust-$H_{90}$ is an
operational, tail-sensitive development criterion that penalizes large errors in
either fold; it is not a predictive interval, a theoretically optimal objective,
or an F1 objective.

Three similarly named quantities have distinct roles. Development-time
robust-$H_{90}$ selects the frozen specification. The scoring-stability interval,
which governs abstention in the primary test, is a case-bootstrap interval passed
through that frozen proxy. In the supplement's rolling-origin analysis a robust-$H_{90}$ band is re-estimated
from prior experiments only and governs each origin's decisive calls.
Neither band is the scoring-stability interval.

\paragraph{Scoring-stability intervals.}
We pass bootstrap draws of the offline inputs through the fixed fitted map:
\[
  I^{\mathrm{offline}}_i =
  \operatorname{Quantile}_{2.5,97.5}\!\left(
    \left\{f_{\widehat{\theta}_{\widehat{s}}}
    \!\left(\phi_{\widehat{s}}(x_{i,\mathrm{EC}}^{(b)},
    x_{i,\mathrm{aux}})\right)\right\}_{b=1}^{10{,}000}
  \right).
\]
Each contrast pairs control and treatment by case ID and draws 10{,}000 samples
with replacement under a fixed seed, applying the same indices to both arms. It
resamples retained EC scores rather than rerunning generation or the scorer, and
auxiliary features stay fixed, so the interval excludes generation and scorer
randomness, calibration-fit and auxiliary-feature uncertainty, and future online
residual variation. The frozen export is stored in its native
three-day scale and joined to seven-day online outcomes;
\S\ref{sec:robustness} reports that join's horizon sensitivity. Sign calls and
ranking are invariant to a common positive rescaling but magnitude is not, so we
claim neither that the export is a seven-day effect size nor that the six-hour
training construct is equivalent to either online outcome.

\section{Empirical Study}
\label{sec:empirical}

The study asks whether the frozen proxy agrees with randomized outcomes on
cohorts it could not have seen (\S\ref{sec:postfreeze}), whether it beats the
decision rules a team would otherwise use (\S\ref{sec:baselines}), which
components carry the signal (\S\ref{sec:components}), whether agreement is stable
across horizon and confidence level
(\S\ref{sec:robustness}), and how the suite itself should be
built (\S\ref{sec:scaling}). Only \S\ref{sec:postfreeze} is primary evidence;
Sections~\ref{sec:baselines}--\ref{sec:system} are supporting or exploratory
analyses on slices that overlap the parent corpus, are tuned on the same rows
they are scored on, or come from a different artifact.

\subsection{Evidence Map and Protocol Boundaries}
\label{sec:protocol}
We keep experiments intact and use distinct slices for distinct questions; the
supplement indexes every analysis with its size and what it may be used to
conclude. The primary discovery (90
contrasts, five experiments) and replication (113 contrasts, eight experiments)
cohorts are chronologically disjoint from each other and from the earlier 286-row
completed overlap; the remaining rows overlap that parent corpus, tune on their own
slice, or use a different artifact, and none of them is an independent
replication. Whole-experiment resampling throughout preserves observed cluster
membership, but its percentile ranges are not future-experiment coverage
intervals. All results condition on completed offline and online evaluations
rather than on the full candidate funnel.

\begin{figure}[t]
  \centering
  \includegraphics[width=\columnwidth]{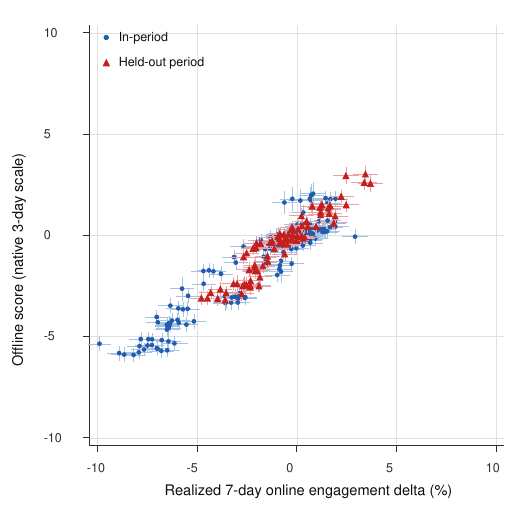}
  \caption{Calibrated offline score in its native three-day scale versus the
  realized seven-day online engagement delta on the completed overlap. The
  descriptive, in-sample, experiment-balanced association is $r=0.961$
  (experiment-resampled 95\% percentile range $[0.939,0.977]$). This slice
  includes development experiments, so the figure illustrates the paired-corpus
  structure rather than providing independent validation; the differing scales
  preclude any seven-day magnitude-calibration reading.}
  \label{fig:calibration_scatter}
\end{figure}

For context only, the descriptive completed overlap reaches 82.9\% contrast-micro
and 86.2\% experiment-macro F1; an experiment-disjoint frozen holdout within that
overlap reaches 87.3\%. Because these slices include development data or overlap
the parent corpus, they support consistency but not the primary replication claim.
Figure~\ref{fig:calibration_scatter} shows the corresponding paired observations.

\subsection{Primary Test: Sequential Post-Freeze Discovery and Replication}
\label{sec:postfreeze}
One versioned specification is held fixed across two chronologically sequential
cohorts. The discovery cohort is every contrast in the first batch of five
experiments that ran after the freeze and was absent from the retained export:
90 in total. It exposed an observed cross-tier raw-score inversion. The
replication cohort is the next batch, read from a frozen copy of the production
data: all 113 eligible seven-day contrasts from eight later experiments. The same
serialized map is applied through the production evaluation pipeline, without
refitting, row selection, or threshold tuning. The offline evaluations were completed before
treatment exposure; we assembled the reports from that copy afterwards. Offline
evaluator inputs exclude online outcomes and treatment-arm interaction logs.
Raw and frozen reports have identical exhaustive keys
$(\text{experiment},\text{control arm},\text{test arm})$.

\begin{table*}[t]
  \centering
  \caption{Sequential post-freeze results. Calls are interval-decisive offline
  predictions; online counts significant online outcomes. ``Wrong'' counts
  observed wrong-direction decisive calls, not total prediction error.}
  \label{tab:postfreeze}
  \tiny
  \setlength{\tabcolsep}{2.4pt}
  \begin{tabular}{lllrrrrrrrr}
    \toprule
    Cohort & Stratum & Score & $N$ & Calls & Online & Correct & Wrong & Prec. & Rec. & F1 \\
    \midrule
    Discovery & All & Raw EC & 90 & 32 & 23 & 0 & 22 & 0.0 & 0.0 & 0.0 \\
    Discovery & All & Frozen composite & 90 & 34 & 23 & 22 & 0 & 64.7 & 95.7 & 77.2 \\
    \midrule
    Replication & All & Raw EC & 113 & 72 & 68 & 24 & 31 & 33.3 & 35.3 & 34.3 \\
    Replication & All & Frozen composite & 113 & 75 & 68 & 58 & 0 & 77.3 & 85.3 & 81.1 \\
    Replication & Same-tier & Raw EC & 78 & 40 & 33 & 23 & 0 & 57.5 & 69.7 & 63.0 \\
    Replication & Same-tier & Frozen composite & 78 & 42 & 33 & 25 & 0 & 59.5 & 75.8 & 66.7 \\
    Replication & Cross-tier & Raw EC & 35 & 32 & 35 & 1 & 31 & 3.1 & 2.9 & 3.0 \\
    Replication & Cross-tier & Frozen composite & 35 & 33 & 35 & 33 & 0 & 100.0 & 94.3 & 97.1 \\
    \bottomrule
  \end{tabular}
\end{table*}

\begin{table}[t]
  \centering
  \caption{Replication results for the five multi-contrast experiments. R denotes
  replication; frozen wrong-direction calls are zero in every row. Three further
  experiments (R6--R8) contribute one contrast each and are omitted here; their F1 values of 100.0, 0.0, and 0.0 are what pull
  experiment-macro F1 (63.0\%) below the 80.8\% mean of the rows below.}
  \label{tab:replication_by_experiment}
  \small
  \begin{tabular}{lrrrr}
    \toprule
    Exp. & $N$ & Raw F1 & Frozen F1 & Raw wrong \\
    \midrule
    R1 & 36 & 0.0 & 51.6 & 8 \\
    R2 & 28 & 72.7 & 93.6 & 3 \\
    R3 & 21 & 46.7 & 93.8 & 6 \\
    R4 & 15 & 8.7 & 76.2 & 9 \\
    R5 & 10 & 0.0 & 88.9 & 4 \\
    \bottomrule
  \end{tabular}
\end{table}

The replication (Table~\ref{tab:postfreeze}) repeats the main pattern: contrast-micro F1 is
81.1\% frozen versus 34.3\% raw, experiment-macro F1 is 63.0\% versus 16.0\%,
and observed wrong-direction calls fall from 31 to zero. The correction is not
carried by one cluster: the frozen composite improves on raw in all five
multi-contrast experiments, by between 20.9 and 88.9 points, and makes no
wrong-direction call in any of the eight
(Table~\ref{tab:replication_by_experiment}). The
cross-tier stratum repeats the initially discovered correction (97.1\% frozen F1
versus 3.0\% raw). The larger same-tier stratum shows moderate observed
agreement: frozen F1 is 66.7\% with zero observed wrong-direction calls.
Its incremental 3.7-point gain over raw is not established, however; across
10{,}000 paired whole-experiment bootstrap resamples, frozen-minus-raw F1 has a
95\% percentile range of $[0,+11.1]$ points. For all 113 rows, the corresponding
range is $[+30.8,+68.3]$, and frozen leave-one-experiment-out F1 ranges from
75.0\% to 89.3\%. These cluster sensitivities preserve experiment membership but
are not calibrated future-experiment coverage intervals.

The discovery cohort localizes the failure. All 22 cross-tier contrasts are
online significant, all 22 decisive raw calls have the wrong sign, and the frozen
composite corrects all 22. Reusing the full model's interval widths, a tier-only ablation
reaches 97.8\% F1 while removing tier reproduces the inversion, identifying the
already-frozen tier term as the directional mechanism---though not a deployable
tier-only model, since a tier-only band estimated from the 286 earlier rows would
make zero late calls.

The evidence is cohort-disjoint but not contrast-independent, since contrasts
within an experiment can share controls and populations; neither zero observed
wrong-direction calls nor fixed-point F1 validates confidence calibration, future
coverage, or prospective screening.

\subsection{Comparison Against Naive Rules, Deployed Evaluators, and Reduced Variants}
\label{sec:baselines}
A decision rule is only useful relative to the one it displaces. We therefore
compare the frozen composite against three classes of alternative: rules that
require no model, the offline evaluators that teams actually consult, and reduced
variants of the composite itself
(Table~\ref{tab:baselines_headtohead}).

\begin{table*}[t]
  \centering
  \caption{The frozen composite against less computationally expensive decision rules and against other
  deployed offline evaluators. Panel (b) uses a different artifact and a narrower
  slice than the headline export, so it supports a \emph{relative} ranking of
  evaluators rather than an absolute-F1 comparison with panel (a). Evaluators A–E are other offline evaluators deployed in the same program. $^{*}$A and B use deltas derived
  from OCR-read screenshot net-win-rates and are approximate. Brackets are
  experiment-resampled 95\% percentile ranges (5{,}000 resamples), not nominal
  confidence intervals.  ``Micro''/``Macro'' are contrast-micro and
  experiment-macro F1; ``Wrong'' counts observed wrong-direction calls.}
  \label{tab:baselines_headtohead}
  \footnotesize
  \setlength{\tabcolsep}{3pt}
  \begin{minipage}[t]{0.46\textwidth}
    \centering
    \textbf{(a) less computationally expensive decision rules}\\[2pt]
    \begin{tabular}{lrrr}
      \toprule
      Decision rule & Micro & Macro & Wrong \\
      \midrule
      \multicolumn{4}{l}{\emph{Completed overlap, 286 contrasts / 14 exp.}} \\
      Seeded random sign & 41.0 & -- & 90 \\
      Majority direction & 56.2 & -- & 54 \\
      Raw EC point sign & 71.5 & -- & 18 \\
      Raw EC $+$ abstention & 75.8 & 76.3 & 8 \\
      Frozen composite & 82.9 & 86.2 & 2 \\
      \midrule
      \multicolumn{4}{l}{\emph{Frozen holdout, 84 contrasts / 4 exp.}} \\
      Raw EC $+$ abstention & 70.0 & 74.8 & 6 \\
      Single-signal map & 78.1 & -- & 3 \\
      Frozen composite & 87.3 & 93.1 & 0 \\
      \bottomrule
    \end{tabular}
  \end{minipage}\hfill
  \begin{minipage}[t]{0.50\textwidth}
    \centering
    \textbf{(b) Deployed offline evaluators, shared 157-contrast slice}\\[2pt]
    \begin{tabular}{lrrrr}
      \toprule
      Offline evaluator & Prec. & Rec. & F1 & F1 range \\
      \midrule
      Calibrated EC proxy (ours) & 72.7 & 72.1 & 72.4 & [58.8, 77.5] \\
      Raw EC score & 61.0 & 50.0 & 55.0 & [31.4, 68.7] \\
      Evaluator A$^{*}$ & 44.4 & 49.2 & 46.7 & [29.6, 60.7] \\
      Evaluator B$^{*}$ & 44.3 & 47.5 & 45.8 & [28.4, 61.4] \\
      Evaluator C & 45.2 & 42.6 & 43.9 & [32.5, 48.1] \\
      Evaluator D & 21.0 & 13.9 & 16.7 & [10.5, 26.9] \\
      Evaluator E & 0.0 & 0.0 & 0.0 & [0.0, 0.0] \\
      \bottomrule
    \end{tabular}
  \end{minipage}

  \vspace{5pt}

  \begin{minipage}[t]{\textwidth}
    \centering
    \textbf{(c) Reduced variants of the composite, 84-contrast frozen holdout}\\[2pt]
    \begin{tabular}{lrrr@{\hskip 2em}lrrr}
      \toprule
      Predictor & F1 & Wrong & MAE & Predictor & F1 & Wrong & MAE \\
      \midrule
      Tier only & 48.2 & 0 & 1.719 & EC $+$ tier & 77.7 & 0 & 0.581 \\
      Auxiliary only & 48.2 & 0 & 1.661 & Reduced composite & 82.6 & 0 & 0.539 \\
      EC only & 63.9 & 3 & 1.493 & Frozen full & 87.3 & 0 & 0.559 \\
      \bottomrule
    \end{tabular}
  \end{minipage}
\end{table*}

\paragraph{Less computationally expensive rules.}

On the completed overlap a seeded random-sign rule reaches 41.0\% F1 and an
observed-majority rule 56.2\%; the latter reads its fixed direction from the same
evaluation labels, so it is optimistic by construction and still makes 54
wrong-direction calls. The informative ladder is the rest: raw EC point signs
without intervals reach 71.5\% with 18 wrong-direction calls, adding abstention to
the same score reaches 75.8\% with eight, and the frozen composite reaches 82.9\%
with two. On the frozen holdout a single-signal weighted linear map fit only on
earlier rows reaches 78.1\% with three, versus 87.3\% and zero for the composite
(paired clustered gain $+10.1$, $[7.6,20.0]$). Each added element---abstention,
then multi-signal calibration---removes wrong-direction calls rather than merely
raising F1.

\paragraph{Other deployed evaluators.}
Panel (b) scores seven offline evaluators on a shared slice of 157 contrasts
across seven experiments against the same seven-day outcome. Evaluators A--E are
others deployed in the same program---a mix of model-based judges and rule-based
checkers---with A and B approximate, being derived from OCR-read screenshot
net-win-rates. The calibrated EC proxy reaches 72.4\% F1, exceeding the strongest
alternative by 25.7 points ($[10.7,40.3]$) and raw EC by 17.5 ($[5.4,38.7]$). Raw
EC exceeds the best alternative by only 8.3 points, unresolved at $[-2.3,11.2]$,
so the separation belongs to the composite rather than the classifier alone. This
slice uses each evaluator's deployed per-slice delta, not the calibrated export,
so its absolute F1 is not comparable with the headline numbers.

\paragraph{Reduced variants.}
\label{sec:components}
Refitting reduced versions of the calibration map on the earlier frozen holdout
tests whether the classifier is replaceable by metadata alone. Each component is
individually weak and the combination is not: tier-only and auxiliary-only reach
48.2\% F1, EC-only 63.9\%, EC$+$tier 77.7\%, and the frozen full export 87.3\%,
with MAE falling from 1.719 to 0.559 (panel c). The pattern supports
complementarity rather than an EC-only or metadata-only explanation, but these are
diagnostic refits on one holdout, not a frozen replication ablation, and EC-only
agreement is uneven across the two substantial holdout experiments (80.0\% and 28.6\% F1).

\subsection{Sensitivity to Horizon, Confidence Level, and Clustering}
\label{sec:robustness}
Three design choices could plausibly manufacture the result: the seven-day readout
horizon, the 95\% online confidence level that defines $y_i$, and the decision to
abstain when the offline interval crosses zero. None of them does.

\paragraph{Readout horizon.}
Joining the same matched rows to three-day instead of seven-day readouts, no
contrast that is significant at both horizons changes sign, and online-CI sign
categories agree for 87.8\% of all 286 contrasts and 90.5\% of the 84-contrast
holdout. Directional agreement is thus robust to the horizon, though this does not
establish seven-day magnitude calibration.

\paragraph{Online confidence level.}
Varying the online confidence level over 90\slash 95\slash 99\% moves overall F1 only to
82.4\slash 82.9\slash 81.3\% and holdout F1 to 88.4\slash 87.3\slash 86.0\%. Comparing raw point-estimate
signs instead of intervals gives 83.6\% agreement overall and 88.1\% on the
holdout. The metric remains power-dependent, but it is not an artifact of one
threshold.

\paragraph{Abstention.}
Holding the score fixed at raw EC and only removing interval-based
abstention---forcing a call on every contrast from the point sign---raises
observed wrong-direction calls from eight to 18 on the completed overlap
(Table~\ref{tab:baselines_headtohead}a). Abstention is therefore doing
decision-relevant work rather than inflating a score, and it accounts for a larger
share of the composite's two remaining wrong-direction calls than calibration
does. A geometric partial-credit sensitivity for intervals that cross zero
averages 94.0\% on the completed overlap and 96.6\% on the frozen holdout, and
enters no headline metric.

\paragraph{Experiment clustering.}
Under whole-experiment resampling, composite F1 has a percentile range of
$[76.2,88.0]$ on the completed overlap and $[83.8,100.0]$ on the frozen holdout,
against $[68.0,82.1]$ and $[55.7,80.0]$ for raw EC. Paired, calibration improves
contrast-micro F1 by $+7.3$ points on the overlap ($[0.4,15.2]$) and $+18.7$
points on the holdout ($[11.1,28.8]$); holdout composite F1 stays within
85.2--89.7\% when each experiment is omitted in turn, and per-experiment gains
over raw range from $-10.3$ to $+33.3$ points, so the improvement is not uniform.
With four to eight imbalanced clusters,
these ranges describe dependence on the observed grouping and do not support a
standalone significance claim.

\begin{table*}[!t]
  \centering
  \caption{Evaluation-suite design studies, on 130 contrasts from six
  experiments. Top-$K$ compares critical-density selection with the mean of 20
  same-size random samples; rollout uses the top-1K suite; the turn-selection
  bundles differ in membership and size. F1 and P@20\% are higher-is-better,
  $H_{90}$ and MAE lower-is-better. F1 and P@20\% here follow the CI-aware
  convention of the suite-design analysis rather than the hard rule of
  \S\ref{sec:metric}, so they are not comparable with
  Tables~\ref{tab:postfreeze} and~\ref{tab:baselines_headtohead}. These are
  exploratory, same-slice results.}
  \label{tab:suite_experiments}
  \footnotesize
  \setlength{\tabcolsep}{3.5pt}
  \begin{minipage}[t]{0.48\textwidth}
    \centering
    \textbf{(a) Critical-density top-$K$ vs.\ random}\\[3pt]
    \begin{tabular}{rrrr}
      \toprule
      $K$ & Critical F1 & Random F1 & Lift \\
      \midrule
      100     & 66.3 & 32.4 & $+33.9$ \\
      250     & 70.2 & 49.5 & $+20.7$ \\
      500     & 74.7 & 59.5 & $+15.3$ \\
      1{,}000 & 68.2 & 65.3 & \phantom{0}$+2.9$ \\
      2{,}000 & 79.0 & 72.3 & \phantom{0}$+6.7$ \\
      5{,}000 & 77.6 & 77.6 & \phantom{$+$0}$0.0$ \\
      \bottomrule
    \end{tabular}
  \end{minipage}\hfill
  \begin{minipage}[t]{0.48\textwidth}
    \centering
    \textbf{(b) Multi-turn rollout depth (top-1K)}\\[3pt]
    \begin{tabular}{rrrr}
      \toprule
      Turns & F1 & P@20\% & Cal.\ $H_{90}$ \\
      \midrule
      0 & 68.2 & 72.2 & 1.380 \\
      1 & 79.4 & 72.2 & 1.070 \\
      2 & 79.2 & 72.2 & 1.008 \\
      3 & 81.9 & 88.9 & 1.065 \\
      4 & 81.2 & 94.4 & 1.065 \\
      8 & 76.2 & 77.8 & 1.351 \\
      \bottomrule
    \end{tabular}
  \end{minipage}

  \vspace{8pt}

  \begin{minipage}[t]{0.48\textwidth}
    \centering
    \textbf{(c) Cost-matched scoring budget}\\[3pt]
    \begin{tabular}{rrrrrr}
      \toprule
      Calls & Single F1 & Top-$K$ & Turns & Rollout F1 & Lift \\
      \midrule
      1{,}000 & 68.2 & 250     & 3 & 79.8 & $+11.6$ \\
      2{,}000 & 79.0 & 400     & 4 & 83.9 & \phantom{0}$+5.0$ \\
      5{,}000 & 77.6 & 1{,}000 & 4 & 81.2 & \phantom{0}$+3.6$ \\
      \bottomrule
    \end{tabular}
  \end{minipage}\hfill
  \begin{minipage}[t]{0.48\textwidth}
    \centering
    \textbf{(d) Turn-selection bundle}\\[3pt]
    \begin{tabular}{lrrr}
      \toprule
      Suite & F1 & $H_{90}$ & MAE \\
      \midrule
      Last turn     & 65.3 & 4.68 & 2.02 \\
      Critical turn & 66.2 & 4.37 & 1.97 \\
      \bottomrule
    \end{tabular}
  \end{minipage}
\end{table*}

\subsection{Exploratory Evaluation-Suite Design Studies}
\label{sec:scaling}
Three studies address the Layer-3 design questions raised in
\S\ref{sec:evalsuite-method}---how deep to roll out, how many cases to score, and
which turn to cut at---and are summarized in Table~\ref{tab:suite_experiments}. They tune on their analysis slices and do not
provide independent validation of the frozen proxy.

\subsubsection{Study 1: Multi-turn rollout depth}
On a 130-contrast, six-experiment design slice, the critical-density top-1K suite
improves from 68.2\% F1 at turn 0 to 81.9\% at three simulated turns; four turns
maximize P@20\% at 94.4\%, and deeper rollout degrades. The turn-3/4 elbow is not
an artifact of one suite size: across the top-500, top-1K, top-2K, and top-5K
suites, the calibrated $H_{90}$, calibrated MAE, and Pearson $r$ optima all fall
within turns 2--4, and at turn~3 in most cases. The F1 optimum is the least stable
choice, at turn~4, 3, 1, and 4 respectively.

The practitioner-relevant version of the question is what to buy with a fixed
scoring budget, and here depth dominates breadth
(Table~\ref{tab:suite_experiments}c). At exactly matched budgets of 1{,}000,
2{,}000, and 5{,}000 classifier calls, spending them on continuation rather than
on more turn-0 cases raises F1 from 68.2 to 79.8, 79.0 to 83.9, and 77.6 to 81.2.
 The 1{,}000-call point is the sharpest: 250 critical cases rolled out three turns
reach 79.8\% F1, beating not only the 68.2\% from scoring 1{,}000 cases once but
also the 77.6\% from scoring the entire 5{,}000-case pool---a $5\times$ smaller
suite at one-fifth the scoring cost. Scoring 1{,}000 critical cases through three
turns reaches 81.9\%, likewise above the full-pool 77.6\%.

Two caveats bound this. The rollout strategy at each budget was chosen by F1 on
the same slice, and on a broader matched 218-contrast, 11-experiment lifecycle
slice the static-to-three-turn gain is only $+6.6$ points with an
experiment-clustered range of $[-6.7,23.7]$---directionally positive but
unresolved. Enlarging the rollout suite from top-1K to top-2K on 221 matched
contrasts is approximately null ($+0.9$, $[-3.4,4.7]$).

\subsubsection{Study 2: Top-$K$ critical-density selection}
Against 20 same-size random samples drawn from the same 5{,}000-case pool,
critical-density ranking improves F1 by 33.9 points at $K=100$, 20.7 at $K=250$,
and 15.3 at $K=500$ (Table~\ref{tab:suite_experiments}a). The gain is a
sample-efficiency claim at sub-pool budgets, not universal dominance: it is
nonmonotone, narrows to 2.9 points at $K=1{,}000$, does not improve ranking
precision at every $K$, and vanishes at $K=5{,}000$ where both strategies score
the identical full pool by construction. Because the random baseline samples
within the ranked, loadable pool rather than the original source universe, it is a
conservative comparator for curation but not a test of the pool itself.

Curation and rollout interact, and the interaction reverses the ordering. Scored
once at turn~0, a 1{,}000-case critical suite is worse than the full pool (68.2\%
versus 77.6\% F1). Comparing each suite at its own best depth for each
metric, it is better: top-1K reaches 81.9\% F1 and 94.4\% P@20\%, against 80.2\%
and 77.8\% for the complete 5{,}000-case pool, with top-500 higher still at
82.7\% F1. The pool is given its own optimum too, so the comparison favors it. Because the
critical top-5{,}000 set \emph{is} the pool the random baseline samples from, a
1{,}000-case curated suite with rollout dominates scoring every available case
once. Adding the low-density tail buys coverage but dilutes the between-arm
contrast.

\subsubsection{Study 3: Critical-turn versus last-turn suite bundles}
On 243 historical contrasts, the critical-turn bundle reaches 66.2\% F1 versus
65.3\% for last turn and narrows $H_{90}$ from 4.68 to 4.37, while changing both suite
membership and size. Token-fit filtering removes and refills 801 of 5{,}000
dialogs. These results guide suite construction but are exploratory.

\subsection{Secondary Descriptive Tail Ranking for Human Review}
\label{sec:system}
We evaluate tail ranking on completed experiments as an input to human review, not as evidence
of automatic launch approval, full-funnel screening, or traffic savings. The ranking feeds two decisions: which training checkpoints of a
candidate model justify an experiment slot, and which system-prompt variants to
promote. Both frequently compare configurations drawn from one model tier, and
\S\ref{sec:postfreeze} leaves the composite's incremental advantage over the raw
score unresolved on that stratum, so the cross-tier correction behind the headline
gap should not be assumed to transfer to them. Under the
$k=\min(3,\lceil0.2m\rceil)$ rule, exact ranking matches 19/21 (90.5\%) top-tail
and 15/21 (71.4\%) bottom-tail slots on the completed overlap; on the later
holdout it matches 7/7 (100.0\%) and 6/7 (85.7\%), respectively. When boundary
swaps unresolved by the online confidence
interval are forgiven, the sensitivity rises to 20/21 in each overall tail and
7/7 in each holdout tail. These slot estimates are descriptive and clustered
within experiments.

Ranking by bootstrap margin also yields a usable operating curve. Retaining the
most confident 25\slash 50\slash 75\slash 100\% of offline-significant calls covers
16.4\slash 32.9\slash 49.3\slash 65.4\% of completed contrasts at 100.0\slash 100.0\slash 92.2\slash 82.9\% observed
precision, with 0\slash 0\slash 1\slash 2 wrong-direction calls: a reviewer who acts only on the
most confident half of decisive calls sees no observed sign error on this set.
This completed-set curve does not estimate
screening performance or traffic savings over the full funnel, and it is not a
basis for automatic exclusion (\S\ref{sec:ethics}).

\section{Conclusion}
\label{sec:conclusion}
We introduced a layered checklist---label--outcome, scorer--behavior, and
suite--experiment alignment---and an interval-aware protocol for auditing frozen
offline proxies at the experiment decision boundary. In the disjoint replication
cohort, the frozen composite reaches 81.1\% contrast-micro and 63.0\%
experiment-macro F1, versus 34.3\% and 16.0\% for raw EC, with zero versus 31
observed wrong-direction calls. The effect is strongest cross-tier; on 78 same-tier
rows, moderate observed agreement remains but incremental improvement over raw is
unresolved. The supporting studies agree on where the signal lives: the composite
exceeds the strongest of five other deployed offline evaluators by 25.7 F1 points
on a shared slice, no single component reproduces it in the ablation, abstention
rather than the raw score removes most wrong-direction calls, and the ordering
survives changes to readout horizon and confidence level. On the suite-design
side, spending a fixed scoring budget on interaction depth beats spending it on
more single-response cases at every budget we could match. The contribution is
therefore validation and diagnosis of a composite
decision rule, not an EC-only advance.

In deployment the signal prioritizes candidates for two decisions that generate
more of them than the experiment calendar can absorb: which training checkpoints
justify a slot, and which system-prompt variants to promote. Both largely produce
same-tier comparisons, which is precisely the stratum where the incremental gain
is unresolved---so the operational case rests on the composite's moderate observed
same-tier agreement and its zero wrong-direction calls, not on the cross-tier
correction that drives the headline gap.

Every offline prediction preceded its experiment, but we chose the analysis, including the
tier split, after the outcomes were known rather than fixing it in advance. The completed-experiment sample does not
measure full-funnel screening, traffic savings, future coverage, or subgroup
generalization, and Layers 1--2 remain design requirements rather than independently
validated claims. Engagement is not user welfare; the proxy should inform human
review, not automatic launch decisions. Fixing an analysis plan in advance, broader
same-tier evidence, and per-workflow tier stratification are the next tests.

\section{Ethical Considerations}
\label{sec:ethics}
\paragraph{The optimization target is not user welfare.}
The proxy is trained on, and evaluated against, a session-engagement construct:
whether a user returns within six hours, and the seven-day arm-level engagement
delta. Engagement is a product outcome, not a measure of whether an interaction
was good for the person having it. A system tuned toward it can reward
unnecessarily prolonged interaction, and the six-hour return label in particular
does not distinguish a user who returns because the assistant was useful from one
who returns because it was not. We therefore treat the proxy as an instrument for
prioritizing which candidates receive scarce experiment traffic, never as a launch
criterion or a welfare measure. Randomized experiments and human review remain the
basis for launch decisions.

\paragraph{Screening risk falls asymmetrically on unmeasured groups.}
A proxy used to allocate experiment capacity can discard a candidate that would
have helped users. Our audit reports agreement on completed experiments only, so it
cannot bound how often that happens across the full candidate funnel, and we make
no subgroup claims: the corpus is not stratified by language, locale, or user
population, so a candidate that helps an underrepresented group while moving
aggregate engagement little could be screened out without appearing in any metric
we report. The abstention mechanism partially mitigates this by declining to call
contrasts whose offline interval crosses zero, and we recommend that the bottom
tail route candidates to human review rather than to automatic rejection.

\paragraph{Data provenance.}
The evaluation suite is built from logged production conversations. Records are deduplicated, and filtered to remove templated and
low-quality content; no raw conversation text appears in this paper. All experiment identifiers, arm labels, model tiers, and calendar
dates are masked, and the reported quantities are aggregate statistics over
contrasts rather than per-user measurements. The user-simulation extension
generates synthetic continuations rather than replaying real user turns. The study
analyzes experiments that had already completed; it created no additional data collection.

\paragraph{Scope of the claims.}
The evidence supports a frozen composite decision rule for prioritization on one
assistant in one product setting. It does not establish
calibrated future coverage, full-funnel screening performance, traffic savings, or
generalization to other products or engagement definitions. We have tried to state
these boundaries wherever a number is reported rather than only in the limitations
paragraph, because an offline proxy that is trusted beyond its validated range is
precisely the failure mode this paper is arguing against.

\bibliographystyle{ACM-Reference-Format}
\bibliography{references}

\clearpage
\appendix
\section*{Supplementary Material}
\addcontentsline{toc}{section}{Supplementary Material}
\section{Evidence Accounting}
\label{sec:evidence_accounting_supp}
Table~\ref{tab:evidence_map_supp} indexes every analysis in the paper and in this
document: its size, and what it may be used to conclude. Only
the two post-freeze cohorts form the primary sequential test; the remaining slices
are supporting, robustness, or exploratory evidence. Throughout this document, EC
denotes the engagement classifier defined in the main paper, and a
wrong-direction call is a decisive offline call whose sign opposes the online
outcome.

\begin{table*}[t]
  \centering
  \caption{Complete evidence map. Rows answer distinct questions and are not
  pooled or treated as independent replications.}
  \label{tab:evidence_map_supp}
  \footnotesize
  \begin{tabular}{lrrp{0.56\textwidth}}
    \toprule
    Analysis & Contrasts & Experiments & Evidentiary status \\
    \midrule
    Completed overlap & 286 & 14 & Descriptive parent slice; includes development experiments. \\
    Frozen later holdout & 84 & 4 & Subset of the completed overlap; experiment-disjoint from calibration selection. \\
    Earlier same-tier & 211 & 13 & Component stratum of the completed overlap; not independent replication. \\
    Post-freeze discovery & 90 & 5 & First disjoint batch; frozen map and offline outputs predate exposure. \\
    Post-freeze replication & 113 & 8 & Subsequent disjoint exhaustive batch; unchanged map and pre-exposure offline outputs. \\
    Calibration search & 552 & 19 & Development-only specification selection. \\
    Nested robustness & 254 & 12 & Related feature-complete universe; 236 rows overlap the completed slice. \\
    Horizon sensitivity & 286 & 14 & Matched three- versus seven-day online-CI signs. \\
    Rollout/top-$K$ design & 130 & 6 & Same-slice exploratory selection. \\
    Rollout cross-check & 96 & 4 & Descriptive out-of-slice check. \\
    Matched rollout & 218 & 11 & Descriptive broader comparison; clustered effect unresolved. \\
    Last- vs.\ critical-turn suite & 243 & 19 & Historical bundles with different membership and size. \\
    Evaluator head-to-head & 157 & 7 & Distinct artifact with two approximate comparator deltas. \\
    Naive-rule baselines & 286 / 84 & 14 / 4 & Same-slice comparators; the majority rule is optimistic by construction. \\
    Component ablation & 84 & 4 & Post hoc refits on earlier rows; not a frozen replication ablation. \\
    Confidence-level sweep & 286 & 14 & Re-analysis of the same rows at 90/95/99\% online confidence. \\
    Computational Cost-matched budget & 130 & 6 & Same-slice; rollout strategy tuned per budget. \\
    \bottomrule
  \end{tabular}
\end{table*}

\section{Sequential Post-Freeze Cohort Details}
\label{sec:postfreeze_supp}

\subsection{Later Live-Snapshot Replication Cohort}
The replication cohort was read from a frozen copy of the production data after
all included experiments completed. Its exhaustive query retains every
eligible seven-day report row in the next post-discovery batch: 113 contrasts from
eight subsequent experiments. Reports are joined on the exact key
$(\text{experiment},\text{control arm},\text{test arm})$; the extraction asserts
equality of the raw and frozen key sets and fails on any mismatch. No experiment
or contrast was dropped, and no coefficient, threshold, or decision band was
refit. The tier strata and analysis were not fixed in advance.

The serialized frozen specification is unchanged. The report invokes the
production evaluation pipeline to apply that artifact, rather than reconstructing
or refitting its map. Tier strata are read from the same seven-day augmented
artifact used by that pipeline. Both the frozen map and the underlying offline
evaluation outputs were completed before treatment exposure. That copy, the report
assembly, and the present analysis all came later. Offline
evaluator inputs exclude treatment logs and online outcomes. This cohort is
disjoint from both the earlier 286-row completed overlap and the 90-row
discovery cohort.

\begin{table*}[t]
  \centering
  \caption{Replication results by anonymized experiment, ordered by cluster size.
  ``Raw wrong'' counts wrong-direction raw calls; the remaining count columns describe
  the frozen composite.}
  \label{tab:replication_experiments}
  \small
  \begin{tabular}{lrrrrrrrr}
    \toprule
    Exp. & $N$ & Raw F1 & Raw wrong & Calls & Online & Correct & Wrong & Frozen F1 \\
    \midrule
    R1 & 36 & 0.0 & 8 & 20 & 11 & 8 & 0 & 51.6 \\
    R2 & 28 & 72.7 & 3 & 25 & 22 & 22 & 0 & 93.6 \\
    R3 & 21 & 46.7 & 6 & 15 & 17 & 15 & 0 & 93.8 \\
    R4 & 15 & 8.7 & 9 & 9 & 12 & 8 & 0 & 76.2 \\
    R5 & 10 & 0.0 & 4 & 4 & 5 & 4 & 0 & 88.9 \\
    R6 & 1 & 0.0 & 1 & 1 & 1 & 1 & 0 & 100.0 \\
    R7 & 1 & 0.0 & 0 & 1 & 0 & 0 & 0 & 0.0 \\
    R8 & 1 & 0.0 & 0 & 0 & 0 & 0 & 0 & 0.0 \\
    \bottomrule
  \end{tabular}
\end{table*}

The eight cluster sizes are 36, 28, 21, 15, 10, 1, 1, and 1. We draw 10{,}000
paired whole-experiment bootstrap samples with replacement using a fixed
deterministic seed, pool
the rows in each draw, and recompute both F1 values. Table~\ref{tab:replication_uncertainty}
also reports leave-one-experiment-out (LOEO) sensitivity. Percentile and LOEO
ranges describe dependence on these observed clusters; they are not nominal
confidence intervals or calibrated future-experiment coverage.

\begin{table*}[t]
  \centering
  \caption{Experiment-clustered replication sensitivities. F1 and ranges are in
  percentage points; $\Delta$ is frozen minus raw.}
  \label{tab:replication_uncertainty}
  \small
  \begin{tabular}{lrrlll}
    \toprule
    Stratum & $N$ & Experiments & Raw F1 range & Frozen F1 range & $\Delta$ range; median \\
    \midrule
    All & 113 & 8 & [1.2, 60.0] & [61.0, 93.0] & [30.8, 68.3]; 47.7 \\
    Same-tier & 78 & 7 & [5.1, 86.2] & [5.1, 88.6] & [0.0, 11.1]; 3.5 \\
    Cross-tier & 35 & 6 & [0.0, 9.8] & [91.9, 100.0] & [88.0, 100.0]; 94.1 \\
    \bottomrule
  \end{tabular}
\end{table*}

Frozen LOEO F1 ranges are 75.0--89.3\% overall, 47.6--83.3\% for same-tier,
and 96.2--100.0\% for cross-tier rows. The same-tier result therefore indicates
moderate observed agreement for the frozen composite, but not an
incremental improvement over raw EC: the paired range includes zero. The much
larger overall and cross-tier differences repeat the mechanism found in the
discovery cohort.

\subsection{Evaluation-versus-Materialization Provenance}
Artifact identifiers were assigned during later import or report assembly.
Numeric suffixes in those identifiers are not authoritative evaluation-execution
timestamps and are therefore not used to infer scoring time or define a timing
subset. The authoritative provenance is that all underlying offline evaluations
and the frozen map predate treatment exposure. The later snapshot extraction and the split by
model tier were both chosen after the outcomes were known.

\subsection{Descriptive Pool Across Both Post-Freeze Cohorts}
Pooling is not used for a replication claim because the cohorts answer sequential
discovery and replication questions. For completeness, their descriptive 203-row
totals are shown in Table~\ref{tab:combined_postfreeze}.

\begin{table*}[t]
  \centering
  \caption{Descriptive totals across the 90-row discovery and 113-row replication
  cohorts.}
  \label{tab:combined_postfreeze}
  \small
  \begin{tabular}{llrrrrrrrr}
    \toprule
    Stratum & Score & $N$ & Calls & Online & Correct & Wrong & Prec. & Rec. & F1 \\
    \midrule
    All & Raw EC & 203 & 104 & 91 & 24 & 53 & 23.1 & 26.4 & 24.6 \\
    All & Frozen composite & 203 & 109 & 91 & 80 & 0 & 73.4 & 87.9 & 80.0 \\
    Same-tier & Raw EC & 146 & 50 & 34 & 23 & 0 & 46.0 & 67.7 & 54.8 \\
    Same-tier & Frozen composite & 146 & 54 & 34 & 25 & 0 & 46.3 & 73.5 & 56.8 \\
    Cross-tier & Raw EC & 57 & 54 & 57 & 1 & 53 & 1.9 & 1.8 & 1.8 \\
    Cross-tier & Frozen composite & 57 & 55 & 57 & 55 & 0 & 100.0 & 96.5 & 98.2 \\
    \bottomrule
  \end{tabular}
\end{table*}

\subsection{Initial Discovery Cohort: Construction and Provenance}
The frozen specification was serialized before all five experiments in the
discovery cohort began. The cohort rule is exhaustive: all 90 contrasts
in the first post-freeze batch absent from the retained frozen export.
No row, experiment, feature,
coefficient, threshold, or decision band was selected or refit using these
outcomes.

The static-suite offline evaluation outputs were completed before experiment exposure.
Formatting-density and output-token features were recovered later from those
immutable original outputs through the official feature extractors, and tier
direction follows the same metadata convention retained in the frozen calibration
data. Applying the official frozen-calibration path to the recovered file produces
90/90 point predictions identical to the independent manual reconstruction
(maximum absolute difference 0). The analysis and its cross-/same-tier strata were
not fixed in advance. It is therefore failure localization on pre-exposure
evidence analyzed after the outcomes were known, not a forward-looking test of a
predeclared plan.

This 90-row cohort is chronologically disjoint from the earlier 286-row completed
overlap. By contrast, the 84-row frozen holdout and 211-row same-tier stratum are
subsets of those 286 rows. The 254-row rolling-origin reduced-feature universe
shares 236 contrasts with the completed overlap and contains 18 additional
feature-complete rows absent from the frozen-export join. These earlier slices are
related analyses, not independent replications.

\subsection{Per-Experiment Results}
The five cluster sizes are severely imbalanced, and multiple cross-tier contrasts
within an experiment share controls or randomized populations. Table~\ref{tab:postfreeze_experiments}
reports the observed experiment-level results without attaching a five-cluster
uncertainty interval.

\begin{table*}[t]
  \centering
  \caption{Frozen post-freeze results by anonymized experiment. ``Raw wrong'' is
  the number of wrong-direction raw calls; remaining columns describe the frozen
  composite.}
  \label{tab:postfreeze_experiments}
  \small
  \begin{tabular}{lrrrrrrrrr}
    \toprule
    Exp. & $N$ & Raw wrong & Calls & Online & Correct & Wrong & Prec. & Rec. & F1 \\
    \midrule
    L1 & 55 & 10 & 22 & 10 & 10 & 0 & 45.5 & 100.0 & 62.5 \\
    L2 & 28 & 7  & 7  & 8  & 7  & 0 & 100.0 & 87.5 & 93.3 \\
    L3 & 3  & 2  & 2  & 2  & 2  & 0 & 100.0 & 100.0 & 100.0 \\
    L4 & 3  & 2  & 2  & 2  & 2  & 0 & 100.0 & 100.0 & 100.0 \\
    L5 & 1  & 1  & 1  & 1  & 1  & 0 & 100.0 & 100.0 & 100.0 \\
    \bottomrule
  \end{tabular}
\end{table*}

\begin{table*}[t]
  \centering
  \caption{Descriptive leave-one-experiment-out sensitivity for the frozen
  post-freeze composite. This five-cluster range is not a confidence interval.}
  \label{tab:postfreeze_loo}
  \small
  \begin{tabular}{lrrrrrrrr}
    \toprule
    Omitted & Remaining $N$ & Calls & Online & Correct & Wrong & Prec. & Rec. & F1 \\
    \midrule
    L1 ($N=55$) & 35 & 12 & 13 & 12 & 0 & 100.0 & 92.3 & 96.0 \\
    L2 ($N=28$) & 62 & 27 & 15 & 15 & 0 & 55.6 & 100.0 & 71.4 \\
    L3 ($N=3$) & 87 & 32 & 21 & 20 & 0 & 62.5 & 95.2 & 75.5 \\
    L4 ($N=3$) & 87 & 32 & 21 & 20 & 0 & 62.5 & 95.2 & 75.5 \\
    L5 ($N=1$) & 89 & 33 & 22 & 21 & 0 & 63.6 & 95.5 & 76.4 \\
    \bottomrule
  \end{tabular}
\end{table*}

Across omissions, frozen contrast-micro F1 ranges from 71.4\% to 96.0\%. The sensitivity
shows dependence on which of five imbalanced clusters is retained; it does not
estimate repeated-sample or future-experiment coverage.

The frozen composite's experiment-macro F1 is 91.2\%, versus 0.0\% for raw EC.
``Zero wrong-direction'' denotes the observed calls in this cohort, not zero error or
a guarantee for future experiments. Likewise, 77.2\% pooled F1 evaluates one fixed
decision rule; the scoring-stability widths omit auxiliary/tier-feature uncertainty,
calibration-fit uncertainty, and future online residuals and therefore do not
validate confidence calibration or predictive coverage.

\subsection{Directional Decomposition Under Inherited Widths}
For EC-only and full-without-tier, the raw EC interval is transformed through the
corresponding frozen terms. For tier-only and full-without-EC, whose independent
case-bootstrap intervals are unavailable, each row's frozen-full scoring-stability
width is held fixed and recentered on the ablated point. Thus the latter rows are
fixed-width directional ablations, not newly fitted uncertainty models.

\begin{table*}[t]
  \centering
  \caption{Post-freeze directional decomposition by stratum under the stated
  inherited-width convention.}
  \label{tab:postfreeze_components_full}
  \tiny
  \setlength{\tabcolsep}{3pt}
  \begin{tabular}{llrrrrrrrr}
    \toprule
    Stratum & Predictor & $N$ & Calls & Online & Correct & Wrong & Prec. & Rec. & F1 \\
    \midrule
    All & EC only & 90 & 32 & 23 & 0 & 22 & 0.0 & 0.0 & 0.0 \\
    All & Full without tier & 90 & 34 & 23 & 0 & 22 & 0.0 & 0.0 & 0.0 \\
    All & Tier only & 90 & 22 & 23 & 22 & 0 & 100.0 & 95.7 & 97.8 \\
    All & Full without EC & 90 & 22 & 23 & 22 & 0 & 100.0 & 95.7 & 97.8 \\
    All & Frozen full & 90 & 34 & 23 & 22 & 0 & 64.7 & 95.7 & 77.2 \\
    \midrule
    Cross-tier & EC only & 22 & 22 & 22 & 0 & 22 & 0.0 & 0.0 & 0.0 \\
    Cross-tier & Full without tier & 22 & 22 & 22 & 0 & 22 & 0.0 & 0.0 & 0.0 \\
    Cross-tier & Tier only & 22 & 22 & 22 & 22 & 0 & 100.0 & 100.0 & 100.0 \\
    Cross-tier & Full without EC & 22 & 22 & 22 & 22 & 0 & 100.0 & 100.0 & 100.0 \\
    Cross-tier & Frozen full & 22 & 22 & 22 & 22 & 0 & 100.0 & 100.0 & 100.0 \\
    \midrule
    Same-tier & EC only & 68 & 10 & 1 & 0 & 0 & 0.0 & 0.0 & 0.0 \\
    Same-tier & Full without tier & 68 & 12 & 1 & 0 & 0 & 0.0 & 0.0 & 0.0 \\
    Same-tier & Tier only & 68 & 0 & 1 & 0 & 0 & 0.0 & 0.0 & 0.0 \\
    Same-tier & Full without EC & 68 & 0 & 1 & 0 & 0 & 0.0 & 0.0 & 0.0 \\
    Same-tier & Frozen full & 68 & 12 & 1 & 0 & 0 & 0.0 & 0.0 & 0.0 \\
    \bottomrule
  \end{tabular}
\end{table*}

Tier direction completely accounts for the recovered cross-tier direction in this
cohort. EC provides no directional gain here and contributes additional same-tier
calls without detecting the lone same-tier mover. The earlier frozen/reduced
component analysis below remains relevant: on that earlier, more mixed cohort EC
improves over metadata-only variants. The two findings show cohort-dependent
component value rather than support for an EC-only headline claim. Although
tier-only F1 (97.8\%) exceeds frozen-full F1 (77.2\%), inherited full-model widths
mean the tier-only row is not an independently estimated deployable uncertainty
model.

\subsection{Prior-Data-Only Tier-Band Sensitivity}
As a post hoc sensitivity, we estimate an independent tier-only decision band
without using any later outcome. On all 286 chronologically earlier completed-
overlap rows, we find the smallest symmetric half-width around the frozen tier-only
point that overlaps at least 90\% of online confidence intervals. The resulting
$H_{90}$ is 3.1937. Every nonzero tier contribution in the discovery cohort has
absolute magnitude 2.6261, so a tier-only interval formed with this prior-data-only
band crosses zero for every discovery row and makes zero calls.

This sensitivity does not change the directional decomposition: the tier term sets
the corrected cross-tier point direction. It changes the actionability conclusion:
the inherited-width 97.8\% F1 is not independently deployable, and decisive calls
come from the full pipeline's uncertainty construction rather than tier direction
alone.

\subsection{Support Shift Toward a Known Failure Tail}
All 22 discovery-cohort cross-tier raw points fall in $[0.434,0.713]$, above the earlier
cross-tier raw-point 90th percentile of 0.163. Table~\ref{tab:postfreeze_support}
shows that this earlier positive tail already exhibits the same directional
failure.

\begin{table*}[t]
  \centering
  \caption{Historical and later support/stratum comparison.}
  \label{tab:postfreeze_support}
  \small
  \begin{tabular}{llrrrrrrrr}
    \toprule
    Slice & Score & $N$ & Calls & Online & Correct & Wrong & Prec. & Rec. & F1 \\
    \midrule
    Earlier cross-tier & Raw EC & 75 & 64 & 74 & 57 & 6 & 89.1 & 77.0 & 82.6 \\
    Earlier cross-tier & Frozen full & 75 & 72 & 74 & 72 & 0 & 100.0 & 97.3 & 98.6 \\
    Earlier same-tier & Raw EC & 211 & 121 & 113 & 84 & 2 & 69.4 & 74.3 & 71.8 \\
    Earlier same-tier & Frozen full & 211 & 115 & 113 & 83 & 2 & 72.2 & 73.5 & 72.8 \\
    Earlier positive tail & Raw EC & 8 & 7 & 7 & 0 & 6 & 0.0 & 0.0 & 0.0 \\
    Earlier positive tail & Frozen full & 8 & 5 & 7 & 5 & 0 & 100.0 & 71.4 & 83.3 \\
    Discovery cross-tier & Raw EC & 22 & 22 & 22 & 0 & 22 & 0.0 & 0.0 & 0.0 \\
    Discovery cross-tier & Frozen full & 22 & 22 & 22 & 22 & 0 & 100.0 & 100.0 & 100.0 \\
    \bottomrule
  \end{tabular}
\end{table*}

The 211 earlier same-tier rows come from 13 experiments with descending cluster
sizes 36, 36, 36, 28, 26, 21, 6, 6, 6, 3, 3, 3, and 1. The three largest
clusters contain 51.2\% of rows and the six largest contain 86.7\%. To preserve
that grouping, we draw 10{,}000 whole-experiment bootstrap samples with replacement
using a fixed deterministic seed, pool all rows from each sampled experiment, and recompute raw
and frozen F1 with paired draws. The 95\% percentile ranges are $[52.7,81.2]$ for
raw and $[54.5,83.1]$ for frozen; frozen-minus-raw has median $+1.37$ points and
range $[-6.5,+9.9]$. The observed same-tier gain is therefore unresolved.

The discovery result is therefore a case-mix/support shift toward a known raw-score
failure regime. Earlier frozen calibration improves cross-tier F1 by 16.0 points
but same-tier F1 by only 1.0 point, while both earlier same-tier variants retain two
wrong-direction calls. The evidence does not show that the tier signal will dominate
in arbitrary future or same-tier cohorts.

\subsection{Alternate Suite Profiles on the Discovery Cohort}
Table~\ref{tab:postfreeze_profiles} evaluates raw profile intervals only; the
frozen composite is not reconstructed for the alternate profiles. The critical
top-1K static suite reduces observed wrong-direction calls from 22 to 3 chiefly by
abstaining (9 calls versus 32) and still makes zero correct decisive calls.
Rollout profiles do not materially recover accuracy.

\begin{table*}[t]
  \centering
  \caption{Raw suite-profile failure comparison on available discovery contrasts.}
  \label{tab:postfreeze_profiles}
  \small
  \begin{tabular}{lrrrrrrr}
    \toprule
    Profile & $N$ & Experiments & Calls & Online & Correct & Wrong & F1 \\
    \midrule
    Static full & 90 & 5 & 32 & 23 & 0 & 22 & 0.0 \\
    Critical top-1K static & 87 & 4 & 9 & 21 & 0 & 3 & 0.0 \\
    Critical top-1K rollout & 90 & 5 & 25 & 23 & 0 & 7 & 0.0 \\
    Critical top-2K rollout & 90 & 5 & 23 & 23 & 1 & 6 & 4.3 \\
    \bottomrule
  \end{tabular}
\end{table*}

\section{Supporting Evaluation-Suite Evidence}
\label{sec:other}
The following results support the design choices behind the proxy but are not the
core evidence: exploratory evaluation-suite studies, a matched prompt-suite lifecycle
cross-check, and lessons learned. The main paper's evidence-slice index summarizes
how each slice below is used.

\subsection{Exploratory Evaluation-Suite Studies: Multi-Turn}
\label{sec:evalset}
We report these studies on a dedicated prompt-set design slice: 23 target
assistant configurations, 5{,}000 ranked cases per configuration, up to eight
simulated user turns, and 130 direct-prompt online contrasts from six experiments
with a 7-day online engagement readout. This slice uses the paper's primary
online window, but it is a design-ablation slice rather than the calibrated
lifecycle holdout; it should therefore be read as evidence about evaluation-suite
construction rather than as a replacement for the end-to-end lifecycle numbers.

For every turn with criticality at least three on a five-point rubric, the judge
records model-discrimination potential, context dependence, evidence quality,
information gain, observability, and whether the turn is make-or-break. We combine
these as
\[
  d_{ct}=q_{ct}\,Q_{ct}\,o_{ct}\,b_{ct},
\]
where $q_{ct}$ is normalized criticality, $Q_{ct}$ combines discrimination,
context, evidence, and information gain, $o_{ct}$ measures observability, and
$b_{ct}$ boosts make-or-break turns. The highest-density prompt is the cutoff,
with ties broken by criticality and then earlier position. Conversation ranking
aggregates the five densest turns and adds bonuses for repeated make-or-break
events; prefixes of that fixed ranking define the top-$K$ suites.

\begin{table*}[t]
  \centering
  \caption{Critical-density top-$K$ selection versus 20 random samples from the
  same 5{,}000-case loadable pool. F1 is online-CI-aware and P@20\% is the
  CI-forgiven ranking sensitivity; $H_{90}$ and MAE are
  calibrated online-delta errors, so reductions are better.}
  \label{tab:promptset_topk}
  \small
  \begin{tabular}{rrrrrrr}
    \toprule
    Top $K$ & Critical F1 & Random F1 & F1 lift & P@20\% lift & $H_{90}$ reduction & MAE reduction \\
    \midrule
    100 & 66.3 & 32.4 & +33.9 & +9.2 & 0.411 & 0.240 \\
    250 & 70.2 & 49.5 & +20.7 & -10.0 & 0.168 & 0.090 \\
    500 & 74.7 & 59.5 & +15.3 & +0.8 & 0.356 & 0.210 \\
    1{,}000 & 68.2 & 65.3 & +2.9 & +0.8 & 0.231 & 0.117 \\
    2{,}000 & 79.0 & 72.3 & +6.7 & -4.7 & 0.318 & 0.119 \\
    5{,}000 & 77.6 & 77.6 & +0.0 & +0.0 & 0.000 & 0.000 \\
    \bottomrule
  \end{tabular}
\end{table*}

Table~\ref{tab:promptset_topk} shows that critical-density ranking improves
sample efficiency, especially at small budgets. At top-500, the ranked suite
improves interval-aware F1 by 15.3 points relative to random samples from the same
pool, lowers calibrated $H_{90}$ by 0.356, and lowers calibrated MAE by 0.210. The
top-5{,}000 row is a design check: the random and critical strategies both score
the entire pool, so the expected difference is zero. Random-sampling entries are
means over 20 draws; because no across-draw or experiment-clustered interval is
reported, the listed lifts are descriptive design comparisons.

\begin{table*}[t]
  \centering
  \caption{Top-1K rollout-depth sweep. Turn 0 is static current-turn scoring;
  turns 1--8 average raw EC scores through that many simulated user turns. $H_{90}$
  and MAE are lower-is-better after one multiplicative calibration scale.}
  \label{tab:promptset_depth}
  \small
  \begin{tabular}{rrrrrr}
    \toprule
    Turns & F1 & P@20\% & Cal. $H_{90}$ & Cal. MAE & Pearson $r$ \\
    \midrule
    0 & 68.2 & 72.2 & 1.380 & 1.025 & 0.804 \\
    1 & 79.4 & 72.2 & 1.070 & 0.854 & 0.855 \\
    2 & 79.2 & 72.2 & 1.008 & 0.834 & 0.866 \\
    3 & 81.9 & 88.9 & 1.065 & 0.804 & 0.876 \\
    4 & 81.2 & 94.4 & 1.065 & 0.817 & 0.869 \\
    8 & 76.2 & 77.8 & 1.351 & 0.999 & 0.812 \\
    \bottomrule
  \end{tabular}
\end{table*}

Rollout is the clearest curation lever (Table~\ref{tab:promptset_depth}). On the
top-1K suite, moving from static
scoring to three-turn rollout raises interval-aware F1 from 68.2\% to 81.9\%, raises
CI-forgiven P@20\% from 72.2\% to 88.9\%, lowers calibrated MAE from 1.025 to 0.804, and raises
Pearson correlation from 0.804 to 0.876. Four turns produce the best top-tail
ranking precision on this slice, while deeper rollout eventually degrades. The
multi-objective operating region is therefore three to four turns, not the
maximum available horizon.

\begin{table*}[t]
  \centering
  \caption{Rollout optima by suite size. Each cell is the depth at which that
  metric is best, with its value. Calibrated $H_{90}$, calibrated MAE, and Pearson
  $r$ optima all fall within turns 2--4 for every suite size, and at turn~3 in
  most cases, so the turn-3/4 elbow is not an artifact of the top-1K default;
  best-F1 depth is the least stable choice. Depths and values here and in
  Table~\ref{tab:promptset_depth} come from the same per-suite depth sweep, so the
  two tables agree cell for cell where they overlap.}
  \label{tab:promptset_perk}
  \small
  \begin{tabular}{lllll}
    \toprule
    Suite & Best F1 & Best P@20\% & Best cal.\ $H_{90}$ & Best Pearson $r$ \\
    \midrule
    Top-500   & turn 4 (82.7) & turn 2 (94.4) & turn 3 (1.032) & turn 4 (0.876) \\
    Top-1K    & turn 3 (81.9) & turn 4 (94.4) & turn 2 (1.008) & turn 3 (0.876) \\
    Top-2K    & turn 1 (80.8) & turn 6 (83.3) & turn 3 (1.019) & turn 3 (0.876) \\
    Top-5K    & turn 4 (80.2) & turn 0 (77.8) & turn 3 (1.171) & turn 3 (0.852) \\
    \bottomrule
  \end{tabular}
\end{table*}

\begin{table*}[t]
  \centering
  \caption{Average-through-turn versus final-turn whole-conversation scoring, by
  suite size. Each entry is the best policy of that type with its F1 and
  calibrated $H_{90}$. Final-turn scoring is stronger at a tuned shallow depth;
  averaging is the more robust default because it preserves earlier evidence as
  depth grows.}
  \label{tab:promptset_traj}
  \small
  \begin{tabular}{lllll}
    \toprule
    Suite & Best average policy & F1 / $H_{90}$ & Best final-turn policy & F1 / $H_{90}$ \\
    \midrule
    Top-500 & avg turns 0--4 & 82.7 / 1.074 & final turn 3 & 79.8 / 0.904 \\
    Top-1K  & avg turns 0--3 & 81.9 / 1.065 & final turn 3 & 83.5 / 1.011 \\
    Top-2K  & avg turns 0--1 & 80.8 / 1.074 & final turn 3 & 82.5 / 0.944 \\
    Top-5K  & avg turns 0--4 & 80.2 / 1.223 & final turn 1 & 80.2 / 1.045 \\
    \bottomrule
  \end{tabular}
\end{table*}

\begin{table*}[t]
  \centering
  \caption{Computational Cost-matched rollout ablation. At each EC-scoring-call budget, the
  single-response baseline spends calls on more cases, while rollout spends calls
  on continuation depth. The rollout row shown is the F1-best strategy at that
  budget on this design slice.}
  \label{tab:promptset_cost}
  \small
  \begin{tabular}{rrrrrrr}
    \toprule
    Budget & Single F1 & Rollout top-$K$ & Rollout turns & Rollout F1 & F1 lift & $H_{90}$ reduction \\
    \midrule
    1{,}000 & 68.2 & 250 & 3 & 79.8 & +11.6 & 0.284 \\
    2{,}000 & 79.0 & 400 & 4 & 83.9 & +5.0 & 0.270 \\
    5{,}000 & 77.6 & 1{,}000 & 4 & 81.2 & +3.6 & 0.481 \\
    \bottomrule
  \end{tabular}
\end{table*}

Across the three exactly matched budgets in Table~\ref{tab:promptset_cost},
spending calls on simulated continuation yields higher F1 and lower calibrated
$H_{90}$ error half-widths than spending the same budget on more turn-0 cases. Because the rollout
strategy at each budget was selected by F1 on this same design slice, this is tuned
design evidence rather than an independent test. We omit the nominal 10{,}000-call
row because the static arm was capped by the 5{,}000-case loadable pool and was
therefore not actually computational cost matched.

Finally, we compared trajectory scoring rules. Averaging response-level scores
through the rollout is the more robust default, but final-turn scoring can be a
strong tuned-depth variant. On the top-1K suite at turn 3, final-turn scoring
improves F1 from 81.9\% to 83.5\%, P@20\% from 88.9\% to 94.4\%, calibrated $H_{90}$
from 1.065 to 1.011, and calibrated MAE from 0.804 to 0.754 relative to
average-through-turn scoring. The advantage is not stable at deeper horizons:
by turn 8, final-turn F1 drops to 57.5\% and calibrated $H_{90}$ widens to 1.970.
For this reason, the rollout analyses use averaged three-to-four-turn trajectories
as their default, with final-turn scoring treated as a tunable sensitivity analysis.

Tables~\ref{tab:promptset_perk} and~\ref{tab:promptset_traj} report the two
sweeps behind these defaults. Two operational caveats accompany deep rollout:
scored-case missingness grows with depth as more trajectories hit the
skipped-length filter, so any maximum-depth policy should report its effective
scored computational cost; and the mechanism labels behind the behavioral read-out are
heuristic rather than human-audited.

Because the top-$K$, rollout-depth, and budget sweeps select an operating region on
the same 130-contrast design slice, the chosen three-to-four-turn setting should be
read as tuned on this slice rather than an independently
validated optimum. The tables do not attach experiment-clustered uncertainty to
these design-slice differences. The matched 96-contrast lifecycle result below is
a descriptive out-of-slice cross-check.

As a lifecycle cross-check on available 7-day rows, the matched top-1K static
versus top-1K three-turn rollout comparison points in the same direction:
interval-aware F1 improves from 71.6\% to 81.7\%, recall improves from 68.6\% to
82.9\%, and CI-forgiven P@20\% improves from 78.6\% to 85.7\% across 96 contrasts
from four experiments. Expanding that lifecycle rollout profile to top-2K does not
help on the same slice (F1 = 77.5\%), consistent with the prompt-set optimization
evidence that adding lower-density tail cases can dilute the
offline-online signal. Because this cross-check has only four experiments and no
clustered interval reported, it supports directional consistency rather than a
resolved rollout effect; the broader matched comparison in the main paper remains
the inferential reference.

\subsection{Historical Critical-Turn Versus Last-Turn Bundle}
A same-scorer comparison uses the same 243 online contrasts from 19 experiments
but changes both suite membership and size. It therefore evaluates historical
suite bundles rather than isolating a causal cutoff effect.

\begin{table}[t]
  \centering
  \caption{Historical critical-turn versus last-turn suite bundles. Lower $H_{90}$ and
  MAE are better.}
  \label{tab:critical_turn}
  \small
  \begin{tabular}{lrrrr}
    \toprule
    Suite & F1 & Prec./Rec. & $H_{90}$ & MAE \\
    \midrule
    Last user turn & 65.3 & 58.8 / 73.5 & 4.68 & 2.02 \\
    Critical turn & 66.2 & 59.3 / 75.0 & 4.37 & 1.97 \\
    \bottomrule
  \end{tabular}
\end{table}

The critical-turn bundle improves F1 by 0.9 points and narrows $H_{90}$ by 0.31 while
using roughly 5{,}000 rather than 7{,}000 cases. This modest result motivates the
bundle operationally but does not establish that critical cutoff selection alone
causes the difference.

Token fit also matters operationally. A proactive filter-and-refill pass removed
801 of 5{,}000 source dialogs, including 280 of the original top-1K dialogs, and
replenished the requested suite sizes from lower ranks. Separately, before this
filter was enforced, 129 distinct classifier-scoring requests exceeded the verified
16{,}384-token runtime input limit. These are distinct counts: the former measures
proactive filtering, whereas the latter records observed pre-filter runtime
failures. They are not headline
predictive-validity metrics, but they explain why evaluation-suite design must be
reported as part of the EC evaluator rather than treated as an implementation
detail.

\subsection{Earlier Broad and Frozen-Holdout Details}
The earlier experiment-disjoint holdout remains supporting consistency evidence.
It contains two substantial experiments and two very small experiments:

\begin{table*}[t]
  \centering
  \caption{Supporting hard-rule offline-to-online agreement. Brackets are
  experiment-resampled percentile ranges; tail entries are exact matches/slots.}
  \label{tab:fidelity_supp}
  \tiny
  \begin{tabular}{llrrrrrrr}
    \toprule
    Predictor & Split & Prec. & Rec. & F1 & F1 range & Wrong/N & Top & Bottom \\
    \midrule
    Composite proxy & Overall (descriptive) & 82.9 & 82.9 & 82.9 & [76.2, 88.0] & 2/286 & 19/21 & 15/21 \\
    Composite proxy & Frozen holdout & 85.9 & 88.7 & 87.3 & [83.8, 100.0] & 0/84 & 7/7 & 6/7 \\
    Composite proxy & In-period reference & 81.3 & 80.0 & 80.6 & -- & 2/202 & -- & -- \\
    Raw static score & Overall & 76.2 & 75.4 & 75.8 & [68.0, 82.1] & 8/286 & -- & -- \\
    Raw static score & Holdout & 72.4 & 67.7 & 70.0 & [55.7, 80.0] & 6/84 & -- & -- \\
    Majority direction & Overall & 46.5 & 71.1 & 56.2 & -- & 54/286 & -- & -- \\
    Raw point sign & Overall & 59.1 & 90.4 & 71.5 & -- & 18/286 & -- & -- \\
    Seeded random sign & Overall & 33.9 & 51.9 & 41.0 & -- & 90/286 & -- & -- \\
    \bottomrule
  \end{tabular}
\end{table*}

\begin{table}[t]
  \centering
  \caption{Anonymized earlier frozen-holdout results by experiment.}
  \label{tab:holdout_experiments}
  \small
  \begin{tabular}{lrrrrr}
    \toprule
    Exp. & $N$ & Online calls & Proxy calls & Proxy F1 & Raw F1 \\
    \midrule
    E1 & 45 & 35 & 37 & 88.9 & 77.3 \\
    E2 & 36 & 24 & 24 & 83.3 & 55.0 \\
    E3 & 2 & 2 & 2 & 100.0 & 66.7 \\
    E4 & 1 & 1 & 1 & 100.0 & 100.0 \\
    \bottomrule
  \end{tabular}
\end{table}

Under whole-experiment resampling, calibration improves contrast-micro F1 by
$+18.7$ points on this holdout (95\% percentile range $[11.1,28.8]$) and $+7.3$
on the completed overlap ($[0.4,15.2]$). Holdout composite F1 remains
85.2--89.7\% when each experiment is omitted in turn; its gain over raw ranges
from 12.2 to 27.4 points. Composite-proxy macro F1 is 86.2\% overall
($[80.9,91.2]$) and 93.1\% on the holdout ($[86.1,100.0]$). Per-experiment gains
over raw range from $-10.3$ to $+33.3$ points. Overall F1 remains 81.3--82.9\%
when the online confidence threshold varies from 90\% to 99\%. These are
sensitivity ranges, not nominal uncertainty claims for four future clusters.

The paired observations behind these numbers are plotted in the main paper. Their
descriptive in-sample experiment-balanced association is $r=0.961$
(experiment-resampled 95\% percentile range $[0.939,0.977]$), computed on the
completed overlap between the calibrated offline score in its native 3-day scale
and the realized 7-day online engagement delta.

\subsection{Historical Calibration Feature-Family Diagnostic}
A tuning diagnostic compares a classifier-signal-only calibration family with a
family drawn from the pooled pre-exposure features described in the main paper.
The classifier-only family has $H_{80}/H_{90}/H_{95}$ half-widths of
$0.958/2.087/2.909$, while the pooled family has
$1.299/1.735/1.922$ on the native calibration scale. The pooled family therefore
trades a wider central $H_{80}$ band for narrower $H_{90}$ and $H_{95}$ tails,
consistent with robust-$H_{90}$ selection. This is a historical tuning diagnostic,
not a held-out ablation. Exact feature terms
are summarized in the main paper; coefficient values remain omitted.

\subsection{Post Hoc Frozen Component Details}
The component analysis in the main paper was conducted after the frozen proxy was
evaluated; it is a post hoc robustness analysis, not a prospectively specified
ablation. Aggregate results are:

\begin{table*}[t]
  \centering
  \caption{Post hoc frozen-holdout component robustness analysis. Reduced families
  are refit on the earlier 202 rows; ``frozen full'' is the retained export. Calls
  are predicted decisive / 62 online-significant contrasts.}
  \label{tab:components}
  \small
  \begin{tabular}{lrrrrrr}
    \toprule
    Predictor family & Prec. & Rec. & F1 & Calls & Wrong & MAE \\
    \midrule
    Tier only & 95.2 & 32.3 & 48.2 & 21/62 & 0 & 1.719 \\
    Auxiliary only (format + tier) & 95.2 & 32.3 & 48.2 & 21/62 & 0 & 1.661 \\
    EC only & 88.6 & 50.0 & 63.9 & 35/62 & 3 & 1.493 \\
    EC + tier & 97.6 & 64.5 & 77.7 & 41/62 & 0 & 0.581 \\
    Reduced EC + format + tier & 95.7 & 72.6 & 82.6 & 47/62 & 0 & 0.539 \\
    Frozen full & 85.9 & 88.7 & 87.3 & 64/62 & 0 & 0.559 \\
    \bottomrule
  \end{tabular}
\end{table*}

Within this retained reduced family, EC-only reaches 63.9\% F1 versus 48.2\%
for tier-only and auxiliary-only, while EC+tier reaches 77.7\% and the reduced
composite 82.6\%. These diagnostic refits support complementarity rather than an
EC-only or metadata-only explanation. Performance remains uneven across the two
substantial holdout experiments:

\begin{table*}[t]
  \centering
  \caption{Post hoc component agreement on the two substantial frozen-holdout
  experiments. E1 has 45 contrasts and E2 has 36. P/R/F1 are percentages;
  W is the number of wrong-direction calls.}
  \label{tab:component_by_experiment}
  \tiny
  \setlength{\tabcolsep}{3pt}
  \begin{tabular}{lrrrrrrrr}
    \toprule
    Predictor family & E1 P & E1 R & E1 F1 & E1 W & E2 P & E2 R & E2 F1 & E2 W \\
    \midrule
    EC only & 86.7 & 74.3 & 80.0 & 3 & 100.0 & 16.7 & 28.6 & 0 \\
    EC + tier & 96.6 & 80.0 & 87.5 & 0 & 100.0 & 41.7 & 58.8 & 0 \\
    Reduced EC + format + tier & 93.9 & 88.6 & 91.2 & 0 & 100.0 & 50.0 & 66.7 & 0 \\
    Frozen full & 86.5 & 91.4 & 88.9 & 0 & 83.3 & 83.3 & 83.3 & 0 \\
    \bottomrule
  \end{tabular}
\end{table*}

The EC-only result therefore does not support a uniformly strong scorer claim;
the tested composite signals are complementary. A separate legacy single-signal
baseline reaches 78.1\% F1. That baseline is a weighted linear map with an
intercept whose raw interval endpoints are transformed by the fixed map. The
63.9\% EC-only row in the main component table instead uses the common reduced-
family protocol: no-intercept weighted least squares and a newly selected earlier-
data $H_{90}$. The two values are not directly comparable.

\subsection{Post Hoc Fully Nested Reduced-Family Robustness Analysis}
A supplementary expanding-window analysis selects among seven reduced candidate
families assembled post hoc from locally retained features. Selection, fitting,
and rolling-origin robust-$H_{90}$ band estimation use only experiments strictly earlier than each
origin, but the candidate universe itself was not prospectively specified. Across
12 origins and 254 contrasts, pooled precision/recall/F1 is
82.8\slash 64.9\slash 72.8\%, with 14 wrong-direction calls; per-origin F1 has median 85.7\%
and range 35.3--100.0\%. This post hoc robustness analysis selects by minimizing
the maximum of in-sample and leave-one-experiment-out $H_{90}$ on prior data, with
ties broken by fewer features and then family name; it does not select by hindsight
F1.

\begin{table*}[t]
  \centering
  \caption{Same-slice rolling-origin reduced-family baselines. ``Selected'' uses
  the prior-only robust-$H_{90}$ criterion; other rows fix one family throughout.}
  \label{tab:nested_family_baselines}
  \small
  \begin{tabular}{lrrrr}
    \toprule
    Family & Prec. & Rec. & F1 & Wrong \\
    \midrule
    Prior-only selected family & 82.8 & 64.9 & 72.8 & 14 \\
    Auxiliary only, fixed & 79.6 & 48.0 & 59.9 & 12 \\
    EC only, fixed & 88.0 & 51.5 & 64.9 & 4 \\
    EC + tier, fixed & 96.3 & 61.4 & 75.0 & 1 \\
    Reduced full, fixed & 90.9 & 70.2 & 79.2 & 4 \\
    \bottomrule
  \end{tabular}
\end{table*}

Table~\ref{tab:nested_origins} exposes every rolling origin. Reduced full is
selected at nine origins, auxiliary only at two, and tier only at one; each of the
other four families is selected zero times. These counts describe the post hoc
selection path rather than adding an uncertainty analysis.

\begin{table*}[t]
  \centering
  \caption{Per-origin post hoc reduced-family backtest. Origin labels mask dates;
  $n$ is contrast count, ``Prior'' is the number of earlier experiments, and
  ``Pred.''/``Online'' give predicted decisive and online-significant calls.}
  \label{tab:nested_origins}
  \tiny
  \setlength{\tabcolsep}{3pt}
  \begin{tabular}{lrrlrrrrrr}
    \toprule
    Origin & $n$ & Prior & Selected family & Pred. & Online & Prec. & Rec. & F1 & Wrong \\
    \midrule
    O1  & 45 & 4  & Reduced full   & 40 & 35 & 80.0  & 91.4 & 85.3 & 3 \\
    O2  & 10 & 5  & Auxiliary only & 4  & 4  & 100.0 & 100.0 & 100.0 & 0 \\
    O3  & 45 & 6  & Auxiliary only & 23 & 28 & 39.1  & 32.1 & 35.3 & 11 \\
    O4  & 36 & 7  & Tier only      & 8  & 21 & 100.0 & 38.1 & 55.2 & 0 \\
    O5  & 6  & 8  & Reduced full   & 3  & 4  & 100.0 & 75.0 & 85.7 & 0 \\
    O6  & 10 & 9  & Reduced full   & 4  & 5  & 100.0 & 80.0 & 88.9 & 0 \\
    O7  & 6  & 10 & Reduced full   & 3  & 3  & 100.0 & 100.0 & 100.0 & 0 \\
    O8  & 6  & 11 & Reduced full   & 3  & 5  & 100.0 & 60.0 & 75.0 & 0 \\
    O9  & 1  & 12 & Reduced full   & 1  & 1  & 100.0 & 100.0 & 100.0 & 0 \\
    O10 & 8  & 13 & Reduced full   & 3  & 6  & 100.0 & 50.0 & 66.7 & 0 \\
    O11 & 45 & 14 & Reduced full   & 31 & 35 & 96.8  & 85.7 & 90.9 & 0 \\
    O12 & 36 & 15 & Reduced full   & 11 & 24 & 100.0 & 45.8 & 62.9 & 0 \\
    \bottomrule
  \end{tabular}
\end{table*}

Within the common reduced universe, committing globally after inspection to the
reduced-full family reaches 79.2\% F1 versus 72.8\% for prior-only family
selection. This 6.4-point difference combines limited data at early origins,
selection variance, and the advantage of a global commitment; it is not a pure
estimate of selection bias. The separate globally fixed production-family
reconstruction reaches 82.0\% F1. Its residual 2.8-point advantage over the
reduced-full row reflects feature and implementation differences, so it is an
operational reference rather than a same-universe ablation. Neither analysis is a
nested replay of the original search: the retained local snapshot lacks the
output-token feature and several other row-level features needed to reconstruct
all 32 specifications. Consequently, exact original-family nesting is
unrecoverable from this snapshot, and these results are transparent post hoc
robustness analyses.

\subsection{Power-Aware Precision}
The support score $s_i$ is defined in the main paper. Averaging it over decisive
offline calls is exactly the power-aware precision defined there, and it gives
94.0\% on the completed overlap and 96.6\% on the frozen later holdout, against
hard precision of 82.9\% and 85.9\%. The roughly eleven-point gap is the share of
decisive calls whose online interval moves in the predicted direction without
reaching significance---agreement the hard rule scores as zero. Power-aware
precision is a sensitivity analysis, not a probability of correctness, and it
enters none of the recall, F1, or exact-ranking figures, which all use the hard
rule.

\subsection{Head-to-Head Comparison Against Other Offline Evaluators}
\label{sec:headtohead}
The main paper reports this comparison. Seven offline evaluators are scored on a
shared slice of 157 pairwise contrasts across seven experiments, each evaluator's
reported offline delta and interval measured against the same 7-day online session
outcome with the interval-aware significant-mover metric. Experiment-resampled
percentile ranges use 5{,}000 resamples.

Three properties of the slice bound its interpretation. First, it uses each
evaluator's deployed per-slice reported delta rather than the stability-interval
calibrated export, so absolute F1 is not comparable to the headline numbers.
Second, it is narrower than and distinct from the calibrated lifecycle overlap.
Third, evaluators A and B derive their deltas from OCR-read screenshot
net-win-rates and are therefore approximate. The comparison supports a
\emph{relative} ranking of offline evaluators, not an independent estimate of
absolute predictive validity.

\subsection{Lessons Learned}
\label{sec:lessons}
Three lessons are clear. First, the offline evaluation suite is part of the
measurement instrument. For multi-turn assistants, rollout construction,
critical-turn selection, and token-fit handling can materially change what the EC
observes. Second, calibration should be evaluated as a decision procedure with
intervals, not only as a regression fit; in the current 7-day overlap, the
scoring-stability interval changes both recall and wrong-direction behavior relative to
raw classifier-score intervals. Third, ranking metrics are more actionable than correlation
alone when the operational question is which candidates should receive scarce
online experiment slots.

\end{document}